\documentclass{article}
\usepackage[preprint]{neurips_2025}
\usepackage{graphicx}
\usepackage{amsmath,amssymb}
\usepackage{algorithm}
\usepackage{algorithmic}
\usepackage{booktabs}
\usepackage{natbib}
\usepackage{makecell}
\usepackage[utf8]{inputenc} %
\usepackage[T1]{fontenc}    %
\usepackage{hyperref}       %
\usepackage{url}            %
\usepackage{booktabs}       %
\usepackage{amsfonts}       %
\usepackage{nicefrac}       %
\usepackage{microtype}      %
\usepackage{xcolor}         %
\usepackage{amsthm}
\usepackage{enumitem}
\newtheorem{proposition}{Proposition}

\usepackage{multirow}
\usepackage{booktabs}
\usepackage{subcaption}  %
\usepackage{wrapfig}   %

\usepackage{cleveref}
\usepackage{amsmath,amssymb,amsthm}

\newtheorem{theorem}{Theorem}

\numberwithin{theorem}{section}
\numberwithin{proposition}{section}

\theoremstyle{definition}

\usepackage{cleveref}

\let\cite\citep

\title{NdLinear: Preserving Multi-Dimensional Structure for Parameter-Efficient Neural Networks}

\author{
  Alex Reneau\quad Jerry Yao-Chieh Hu\quad Zhongfang Zhuang\quad Ting-Chun Liu\quad Xiang He
  \\
  Ensemble AI \\
  \texttt{\{alex,jhu,zhongfang,ting-chun,xiang\}@ensemblecore.ai}
  \And
  Judah Goldfeder \\
  Columbia University \\
  \texttt{jag2396@columbia.edu}
  \And
  Nadav Timor \\
  Weizmann Institute of Science \\
  \texttt{nadav.timor@weizmann.ac.il}
  \And
  Allen Roush \\
  Wand AI/Thoughtworks\\
  \texttt{allen.roush@thoughtworks.ai}
  \And
  Ravid Shwartz-Ziv \\
  NYU \& Wand AI\\
  \texttt{ravid.shwartz.ziv@nyu.edu}
}

\begin{document}

\maketitle

\begin{abstract}

In deep learning, processing multidimensional inputs (e.g., images, medical scans, and time series) is an important task that often requires flattening the inputs. We introduce \emph{NdLinear}, a drop-in replacement for linear layers that operates directly on tensors, requiring no flattening. By applying transformations separately along each dimension, NdLinear preserves native data structure while achieving dramatic parameter reductions, often by orders of magnitude, with minimal memory overhead. We prove NdLinear maintains expressivity through structured Tucker decomposition while preserving VC-dimension scaling. Extensive experiments demonstrate NdLinear's capacity to achieve significant parameter reductions with substantial wall-clock efficiency gains and minimal memory overhead. For instance, our \emph{NdLinear-LoRA} matches or exceeds standard LoRA on language reasoning tasks using up to $9\times$ fewer parameters. Experiments across CNNs, RNNs, Transformers, and MLPs on vision, language, time-series, and tabular tasks consistently demonstrate NdLinear's efficiency gains. While excelling at axis-separable tasks, NdLinear has limitations with entangled spatial interactions. By processing data in its original N-dimensional form, NdLinear provides a theoretically grounded, practical component for building more efficient neural architectures.

\def\thefootnote{}
\footnotetext{NdLinear is licensed by Ensemble AI under the Apache 2.0 license. Our technical contributions and research findings are free for use by the open-source community.}

\end{abstract}

\section{Introduction}

Deep learning excels at processing multidimensional data, such as medical scans, videos and sensor arrays. Yet, a fundamental inefficiency persists throughout modern architectures. Consider what happens when a 32×32×32 voxel tensor reaches a linear layer: it gets flattened into a 32,768-dimensional vector, scrambling the careful spatial organization that earlier layers worked to extract. This pattern repeats everywhere linear layers appear: CNN classification heads flatten spatial feature maps, Transformer MLPs flatten structured token representations, and RNN projections flatten temporal features. Beyond this massive inefficiency, flattening forces networks to waste capacity rediscovering that adjacent voxels are related, that temporal continuity matters, and that feature dimensions have distinct meanings. While convolutions preserve local structure, they cannot replace linear layers for global reasoning, channel mixing, or final predictions. This leaves a critical gap: no existing layer can perform general linear transformations while respecting multidimensional structure.

To address this, we introduce \textbf{NdLinear}, a novel linear layer that operates natively on N-D tensors. Unlike vanilla linear layers, NdLinear processes data in its original N-D form by applying distinct linear transformations sequentially along each tensor dimension. This dimension-wise approach inherently preserves the data's N-D structure.

Preserving data's native organization allows NdLinear to offer compelling advantages: It enhances representational power by maintaining structural integrity and facilitating a more natural information flow. It dramatically reduces parameter counts (often by orders of magnitude vs. flattened layers) and ensures efficient computation without sacrificing performance. These efficiencies lead to faster training/inference and lower memory usage. As a general-purpose layer, NdLinear is a versatile, drop-in replacement for vanilla linear layers in many architectures, benefiting diverse domains by enabling models to better leverage N-D data characteristics.

Our work makes the following primary contributions, substantiated by extensive empirical validation:
\begin{itemize}
    \item We articulate the critical limitations of flattening in conventional linear layers for N-D data.
    \item We introduce and formulate \textbf{NdLinear}, a novel N-D linear layer using sequential, dimension-wise transformations to preserve data structure and achieve substantial parameter efficiency.
    \item We empirically demonstrate through comprehensive evaluations that NdLinear significantly enhances model performance, or matches it with drastically fewer parameters, when replacing vanilla linear layers across diverse architectures (CNNs, RNNs, Transformers, MLPs) and data domains, including vision, language, time-series, and tabular data.
    \item We analyze the significant parameter and computational efficiency gains originating from NdLinear's dimension-wise transformation principle.
\end{itemize}

By demonstrating that preserving N-D structure yields both theoretical and practical advantages,  NdLinear challenges the conventional wisdom that flattening is a necessary evil in neural architectures.

\section{Related Work}

Neural networks process multidimensional data through three main approaches,
each with fundamental limitations:

\paragraph{Flattening-based Methods.}
Standard linear layers reshape N-D tensors to 2D matrices, destroying
spatial and dimensional relationships. This forces networks to relearn
structure from scrambled features, requiring excessive parameters.
For example, a $32 \times 32 \times 32$ tensor needs $\sim10^9$ parameters when flattened, versus $\sim10^3$ if structure were preserved.

\paragraph{Specialized Architectures.}
Convolution layers excel at 2D/3D spatial patterns but become unwieldy beyond
three dimensions. Depthwise separable convolutions \citep{chollet2017}
and axial attention \citep{ho2019axial} reduce complexity through factorization
but remain tied to specific spatial operations rather than general
tensor transformations.

\paragraph{Tensor Decomposition Layers.}
Recent work applies tensor algebra to neural networks. Tensor Contraction
Layers (TCL) \citep{kossaifi2020tensor} use multilinear maps  to compress activations to lower dimensions. Tensor Regression Layers (TRL)
\citep{kossaifi2020tensor} parameterize predictions using Tucker or Tensor-Train formats.
Tensor-Train layers \citep{novikov2015tensorizing} decompose weights for compression.
While efficient, these methods focus on specific tasks, dimensionality
reduction or regression, rather than general-purpose transformation.

\paragraph{Gap in Current Methods.}
Despite extensive work on structured layers, no method provides a true
drop-in replacement for linear layers that: (1) operates directly on N-D
tensors without flattening, (2) supports flexible dimension-wise
transformation (expansion or compression), (3) integrates seamlessly
with modern architectures (bias, normalization, dropout). This gap motivates NdLinear,
which we introduce in Section~\ref{sec:ndlinear_method}.

We provide detailed mathematical comparisons between NdLinear and existing
tensor methods (TCL, TRL, TT decomposition) in Appendix~\ref{app:related2}, demonstrating
how our approach differs fundamentally in design philosophy and implementation,
leading to the superior empirical results shown in Section~\ref{sec:results}.

\section{Linear Transformation Preserving N-dimensional Information}
\label{sec:ndlinear_method}
Vanilla linear layers cannot process input tensors directly. They require transforming the inputs into 2D matrices, destroying the original N-D structure. For an N-D input tensor $X \in \mathbb{R}^{B \times D_1 \times \cdots \times D_N}$
with batch size $B$ and feature dimensions $(D_1, \ldots, D_N)$, standard layers flatten it to
$\mathbb{R}^{B \times \prod_i D_i}$ before applying a linear transformation. NdLinear suggests an alternative approach that transforms $X$ directly to $Y \in \mathbb{R}^{B \times H_1 \times \cdots \times H_N}$ without flattening, where each dimension $D_k$ maps to $H_k$, while preserving the N-D structure throughout the transformation.

The NdLinear transformation processes N-D tensors by applying separate linear transformations along each of their feature dimensions sequentially. This contrasts with vanilla linear layers, which flatten the N-D tensor into a 2D matrix, losing the original multidimensional structure. By operating on each dimension independently while preserving the others, NdLinear retains the inherent structural information of the data throughout the transformation.

Conceptually, for an input tensor $X \in \mathbb{R}^{B \times D_1 \times \cdots \times D_N}$, NdLinear learns $N$ distinct weight matrices $W_1, \ldots, W_N$, where each $W_k \in \mathbb{R}^{D_k \times H_k}$ transforms the $k$-th dimension of the input from its original size $D_k$ to a new size $H_k$, with optional bias vectors $b_k \in \mathbb{R}^{H_k}$ per dimension. The transformation is applied iteratively: the output from transforming dimension $k$ becomes the input for transforming dimension $k+1$. The procedure is detailed in Algorithm~\ref{alg:ndlinear}.

\begin{algorithm}[t]
\caption{NdLinear Transformation}
\label{alg:ndlinear}
\begin{algorithmic} %
\REQUIRE Input tensor $X \in \mathbb{R}^{B \times D_1 \times \cdots \times D_N}$ (batch size $B$, original feature dimensions $D_1, \ldots, D_N$), \\
         Target output feature dimensions $H_1, \ldots, H_N$, \\
         Learnable weight matrices $W_k \in \mathbb{R}^{D_k \times H_k}$ for $k=1, \ldots, N$, \\
         Optional learnable bias vectors $b_k \in \mathbb{R}^{H_k}$ for $k=1, \ldots, N$.
\STATE Let $X_{\text{out}} \leftarrow X$.
\FOR{$k = 1$ to $N$}
    \STATE Let current shape of $X_{\text{out}}$ be $(B, S_1, \ldots, S_N)$, where $S_j = H_j$ for $j < k$, and $S_j = D_j$ else.
    \STATE The $k$-th feature dimension (original size $D_k$, current size $S_k=D_k$) is targeted for transformation.
    \STATE Permute $X_{\text{out}}$ to move its $k$-th feature dimension to the last position. Shape becomes $(B, S_1, \ldots, S_{k-1}, S_{k+1}, \ldots, S_N, D_k)$.
    \STATE Reshape $X_{\text{out}}$ to a 2D matrix $X_{\text{matrix}}$ of shape $\left(B \cdot \prod_{j \neq k} S_j, D_k\right)$.
    \STATE Apply linear transformation: $Y_{\text{matrix}} \leftarrow X_{\text{matrix}}W_k + b_k$. ($Y_{\text{matrix}}$ has shape $\left(B \cdot \prod_{j \neq k} S_j, H_k\right)$).
    \STATE Reshape $Y_{\text{matrix}}$ back to N-D form: $(B, S_1, \ldots, S_{k-1}, S_{k+1}, \ldots, S_N, H_k)$.
    \STATE Permute dimensions to place $H_k$ (the new size of the $k$-th feature dimension) back into the $k$-th feature position. Shape becomes $(B, S_1, \ldots, S_{k-1}, H_k, S_{k+1}, \ldots, S_N)$.
    \STATE Update $X_{\text{out}}$ with the result of this transformation.
\ENDFOR
\RETURN $X_{\text{out}}$, now of shape $\mathbb{R}^{B \times H_1 \times \cdots \times H_N}$.
\end{algorithmic}
\end{algorithm}

In practice, these operations (transposing, reshaping, matrix multiplication, then inverse reshaping and transposing) can be efficiently implemented using standard tensor library functions like \texttt{torch.tensordot} or \texttt{einsum}. The key is that each weight matrix $W_k$ only transforms dimension $D_k$ to $H_k$. This operation modifies all entries along the $k$-th mode, performing the same linear transformation on each mode-$k$ fiber of the tensor.

This sequential application can be expressed using tensor notation as a series of mode-$k$ tensor-matrix products \citep{kolda2009tensor}:
$Y = X \times_1 W_1 \times_2 W_2 \cdots \times_N W_N$
, where each product $X \times_k W_k$ transforms the $k$-th mode of the tensor using matrix $W_k$. (Biases $b_k$ are added after each mode-$k$ product). The intermediate result of $X \times_k W_k$ becomes the input for the $\times_{k+1} W_{k+1}$ product.

\subsection{Preserving The Expressiveness of Vanilla Linear Layers}
\label{sec:expressivity}

The dramatic parameter reduction of NdLinear raises a natural question: does this efficiency sacrifice model expressivity? We show that despite using fewer parameters, NdLinear maintains  sufficient representational capacity.

\textbf{Theoretical Guarantee.} We analyze expressivity through VC-dimension, which measures  a model's capacity to fit arbitrary patterns. Following \citet{bartlett2019nearly}, networks with  $P$ parameters have VC-dimension $\Theta(P \log P)$.

\begin{theorem}[Informal; see Appendix~\ref{app:proofs} for formal statement]
\label{thm:vc}
An NdLinear network with $P_{\text{nd}} = d(a + b + c)$ parameters for tensor dimensions
$(a, b, c)$ and hidden dimension $d$ maintains VC-dimension $\Theta(P_{\text{nd}} \log P_{\text{nd}})$
as $d \to \infty$, matching the scaling of vanilla linear layers with $P_{\text{std}}$ parameters.
\end{theorem}

\textbf{Practical Implication.} While NdLinear uses fewer parameters, the reduction is
polynomial and not exponential in the VC-dimension bound. This theoretical guarantee is
validated empirically: NdLinear often \emph{improves} performance despite parameter reduction,  suggesting the structured factorization acts as beneficial regularization rather than a
limiting constraint.

\textbf{Empirical Evidence: Representation Compression.} To understand why fewer parameters improve
performance, we measured von Neumann entropy of learned representations. On the Radius Bump
task~\ref{app:entropy} across varying difficulties, NdLinear consistently produces
15-30\% lower entropy than parameter-matched dense networks while achieving equal or better
test MSE (Appendix~\ref{app:entropy}). This lower entropy, indicating more compressed
representations, aligns with recent theoretical work showing compressed representations generalize better \citep{skean2025layer}. NdLinear's structured factorization inherently
eliminates redundancy while preserving task-relevant information, explaining why dramatic
parameter reduction enhances rather than hurts performance.

\subsection{Fewer Learnable Parameters And Wall-Clock Speedups}
\label{sec:analysis}

A key advantage of NdLinear is its parameter efficiency. Consider transforming
$X \in \mathbb{R}^{B \times D_1 \times \cdots \times D_N}$ to
$Y \in \mathbb{R}^{B \times H_1 \times \cdots \times H_N}$.

\textbf{Parameter Reduction.} Vanilla linear layers require
$(\prod_{i} D_i) \times (\prod_{i} H_i)$ parameters, which grow exponentially
with dimensionality. NdLinear requires only $\sum_{i=1}^{N}(D_i H_i)$ parameters,
which grow linearly. For example, transforming a $32 \times 32 \times 32$ tensor to
the same size requires $\sim10^9$ parameters for vanilla linear but only $3,072$ for
NdLinear, a reduction of six orders of magnitude.

\textbf{Computational Complexity.} NdLinear's FLOPs scale as
$\mathcal{O}(B \cdot N \cdot D^{N+1})$ for cubic tensors with $D_i = H_i = D$, compared to
$\mathcal{O}(B \cdot D^{2N})$ for vanilla linear layers. This yields order-of-magnitude
speedups that increase with tensor dimensionality. The exact FLOP count is:
\begin{align*}
\text{FLOPs}_{\text{NdLinear}} = B \sum_{k=1}^{N} \left(\prod_{j<k} H_j \cdot \prod_{j>k} D_j \cdot D_k H_k\right).
\end{align*}

\textbf{Memory Efficiency.} While theoretical analysis bounds peak memory overhead at
$1/N$ of baseline, which is typically $\leq 33\%$ for 3D tensors (proof in Appendix~\ref{app:memory}), empirical measurements show much smaller overhead
in practice: only 1.1-2.0\% increased peak memory across diverse architectures, with
training time overhead below 1.6\% (Section~\ref{sec:ablations}).

\subsection{Inductive Bias and Domain Alignment}
\label{sec:bias}

NdLinear's efficiency stems from a strong inductive bias: it assumes dimensions can be
transformed independently. This bias has clear benefits and limitations.

\textbf{The Separability Assumption.} NdLinear implements a rank-1 Tucker decomposition,
assuming the transformation can be factorized across dimensions. Crucially, this structure
persists through ReLU-like activations. For 1-homogeneous activations $\sigma$ (ReLU, GELU):
\begin{align*}
\sigma(T(x)) = \sum_{r=1}^{R} \bigotimes_{k=1}^{m} \sigma(u_r^{(k)}),
\end{align*}
where output rank remains bounded even after nonlinearities. This ensures the separability
bias persists through network depth rather than degrading.

\textbf{Domain Alignment.} NdLinear excels when data dimensions represent independent
factors, such as spectrograms (frequency × time), sensor arrays (sensor × time), and tabular data,
where features have separable effects. However, it struggles with dense cross-dimensional
interactions like XOR patterns, checkerboards, or highly entangled spatial features where
dimensions cannot be meaningfully separated.

\subsection{Quantifying the Trade-off: Dial-a-Bias Experiment}
\label{sec:synthetic}

To precisely characterize when NdLinear's bias helps versus hurts, we designed a controlled
experiment interpolating between separable and entangled patterns.

We create synthetic targets that blend separable and entangled components:
\begin{align*}
y = (1 - \alpha) \cdot f_{\text{separable}}(X) + \alpha \cdot f_{\text{entangled}}(X), \quad \alpha \in [0, 1],
\end{align*}
where $X \in [0,1]^{32 \times 32}$ has i.i.d. uniform entries. The separable component
$f_{\text{separable}}$ aggregates per-axis statistics (row/column means, variances), while
$f_{\text{entangled}}$ requires cross-axis interactions (XOR patterns, checkerboards).

\begin{figure}[t]
\centering
\includegraphics[width=0.65\linewidth]{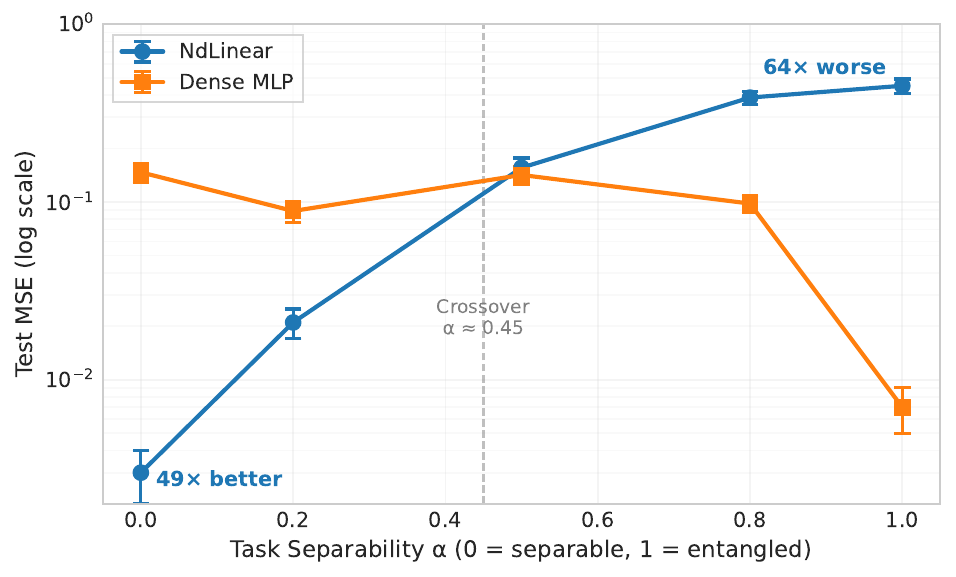}
\caption{\small \textbf{NdLinear excels on separable tasks but struggles with entangled patterns} Performance comparison as task structure varies from purely separable ($\alpha=0$) to fully entangled ($\alpha=1$). The crossover at $\alpha \approx 0.45$ indicates NdLinear outperforms dense MLPs when tasks have $<$45\% entanglement. This provides clear deployment guidance: use NdLinear for axis-aligned domains (spectrograms, time series, tabular data) but prefer dense layers for spatially entangled tasks (dense vision, XOR-like patterns).}

\label{fig:dialabias}
\end{figure}

\textbf{Experimental Protocol.} We trained both NdLinear and dense MLPs (2 hidden layers, 256 units) on 10,000 samples for each $\alpha \in \{0, 0.1, ..., 1.0\}$.

\textbf{Results.}  Figure~\ref{fig:dialabias} reveals a sharp crossover at $\alpha \approx 0.45$.
NdLinear achieves 49× lower MSE for purely separable tasks ($\alpha = 0$) but 64× worse
for fully entangled tasks ($\alpha = 1$). This provides clear guidance on the use of NdLinear, where the target structure aligns with axis-separable patterns.

\section{Experimental Results}
\label{sec:results}

We evaluate NdLinear as a drop-in replacement for vanilla linear layers across language, 
vision, time-series, and tabular domains. Our experiments span CNNs, RNNs, Transformers 
(including LLMs), MLPs, and DiTs, with model scales from 65K to 8B parameters on over 
20 datasets. We also compare NdLinear against alternative structured layers including 
Tensor Regression Layers (TRL/TCL)  and Tensor-Train (TT) decompositions \citep{novikov2015tensorizing, kossaifi2020tensor}

\textbf{Universal findings:} NdLinear reduces parameters by 50-95\% while maintaining or 
improving performance across all tested configurations. When compared to other structured  approaches, NdLinear consistently achieves superior accuracy with lower computational  overhead. We present domain-specific  results below, with ablations in Section~ \ref{sec:ablations} and full details in Appendix~\ref{appen:results}.

\subsection{Natural Language Processing}
\label{sec:nlp_apps}

\subsubsection{Parameter-Efficient Finetuning with NdLinear-LoRA}

We introduce NdLinear-LoRA, replacing LoRA’s low-rank matrices $A,B$ with a single NdLinear adapter. Concretely, the LoRA update $\Delta W = B A$ is swapped for an NdLinear module that applies sequential mode-wise transforms on the reshaped activation tensor; all other LoRA mechanics (zero-init, scaling $\alpha$, residual addition, and merge-at-inference) remain unchanged. This preserves the native $N$-D structure during adaptation while keeping the same drop-in interface and training pipeline as standard LoRA.

We fine-tuned Qwen3-1.7B \citep{yang2025qwen3} and LLaMA3-8B \citep{dubey2024llama3} on mathematical reasoning (OpenMathInstruct-1 \citep{toshniwal2024openmathinstruct}) and commonsense QA  (CommonsenseQA training split \citep{talmor-etal-2019-commonsenseqa}), then evaluated on GSM8K, MultiArith, CSQA, ARC-Easy, ARC-Challenge, and BoolQ..

\textbf{Results.} Table~\ref{tab:ndlinear_lora} shows NdLinear-LoRA achieves superior performance 
with dramatically fewer parameters. On Qwen3-1.7B with 87\% fewer parameters (1.15M vs 8.72M), 
it improves GSM8K by 11.9 points and MultiArith by 7.8 points over standard LoRA. On LLaMA3-8B 
with 9× fewer parameters (2.26M vs 20.97M), it achieves the best CSQA (82.9\%) and ARC-Challenge 
(76.6\%) scores. This suggests structured transformations capture task-specific patterns more  efficiently than low-rank factorization.

\begin{table*}[t]
\centering
\small 
\caption{\small \textbf{NdLinear-LoRA demonstrates significant parameter efficiency for LLM finetuning, achieving comparable or improved accuracy over standard LoRA methods with up to $9\times$ fewer trainable parameters.}
    Accuracy ($\uparrow$) after LoRA of Qwen3-1.7B and LLaMA3-8B models.}%
\label{tab:ndlinear_lora} %
\setlength{\tabcolsep}{4pt} %
\begin{tabular}{@{}llrrrrrrr@{}} 
\toprule
\multicolumn{3}{@{}l}{} & %
\multicolumn{2}{c}{\textbf{Math}} &
\multicolumn{4}{c}{\textbf{CS Reasoning}} \\ %
\cmidrule(lr){4-5} \cmidrule(lr){6-9}
Model & Method & Params %
      & GSM8K & M.Arith %
      & CSQA & ARC‑e & ARC‑c & BoolQ \\
\midrule
\multirow{3}{*}{Qwen3‑1.7B}
  & LoRA ($r{=}4$)   & 4.36M & 45.6 & 88.9 & 80.4 & 91.9 & \textbf{79.4} & 79.7 \\
  & LoRA ($r{=}8$)   & 8.72M & 40.3 & 82.2 & 80.9 & 91.8 & 79.3 & \textbf{80.8} \\
  & NdLinear‑LoRA    & \textbf{1.15M} & \textbf{52.2} & \textbf{90.0} & \textbf{81.0} & \textbf{92.2} & 78.3 & 79.7 \\
\midrule
\multirow{3}{*}{LLaMA3‑8B}
  & LoRA ($r{=}4$)   & 10.48M & 50.5 & \textbf{84.4} & 80.6 & 90.4 & 76.3 & \textbf{85.1} \\
  & LoRA ($r{=}8$)   & 20.97M & \textbf{51.6} & 81.1 & 81.7 & 89.0 & 73.6 & 76.5 \\
  & NdLinear‑LoRA    & \textbf{2.26M} & 40.2 & 80.0 & \textbf{82.9} & \textbf{90.9} & \textbf{76.6} & 80.5 \\
\bottomrule
\end{tabular}

\end{table*}

\subsubsection{Language Model Pretraining}

We pretrained OPT models \citep{zhang2022opt} (124M and 350M parameters) from scratch on BookCorpus and 
Wiki40B-English, replacing feedforward linear layers with NdLinear. Despite fewer 
parameters, NdLinear variants achieve lower perplexity (Table~\ref{tab:linear_vs_ndlinear} in the Appendix). The 
improvement scales with model size: perplexity gap increases from 0.215 (OPT-Small) 
to 0.361 (OPT-Mid), suggesting NdLinear's benefits grow with scale.

\subsubsection{Text Classification}

We evaluated NdLinear in BERT's classification head \citep{devlin2019bert} on SST-2 sentiment analysis \citep{socher2013recursive} and CoLA 
grammatical acceptability \citep{warstadt2018neural} tasks. We replaced BERT's standard two-layer classification head 
with a single NdLinear layer followed by a final linear projection.

\textbf{Results.} As shown in Table~\ref{tab:text_classify} in Appendix~\ref{appen:results}, NdLinear improves both accuracy and 
ROC-AUC on both datasets while reducing the classification head parameters by approximately 
85\%. On SST-2, accuracy improves from 88.72\% to 89.33\%; 
on CoLA, from 77.90\% to 79.06\%.

\subsection{Time Series Analysis}
\label{sec:timeseries_apps}

\textbf{Multivariate Time Series Forecasting.} We evaluated NdLinear in RNNs and Transformers for 12-hour ahead forecasting on four ETT  (Electricity Transformer Temperature) datasets \citep{zhou2021informer}, using 24 hours of historical input. For RNNs, 
we replaced vanilla linear layers with NdLinear. For Transformers, we replaced linear layers  in feedforward blocks with NdLinear.

\textbf{Results.} Table~\ref{tab:ts_forecast_combined_side_by_side} shows NdLinear consistently improves forecasting  accuracy across all datasets while using 50\% fewer parameters. For RNNs on ETTh1, MSE decreases 
from 0.290 to 0.088 (70\% reduction). For Transformers on ETTh2, MSE decreases from 0.0226 to  0.0158 (30\% reduction). These improvements demonstrate NdLinear's natural alignment with  temporal-feature decomposition in multivariate time series.

\textbf{Time Series Classification.} We applied Transformer models with NdLinear to six UCR time series classification datasets \citep{dau2019ucr}, 
replacing vanilla linear layers in transformer blocks with NdLinear layers.

\begin{table*}[t!] %
\centering
\small %
\caption{\small \textbf{NdLinear significantly enhances Transformer-based time series classification, achieving superior F1 scores across UCR datasets with comparable or substantially fewer parameters than baseline models.} F1 Score and efficiency (\# Params) of NdLinear compared to a standard Base Model and a parameter-comparable Base Model$^\star$ (whose parameters are aligned with NdLinear's).
}
\label{tab:ts_tf_classify_final_revised} %
\setlength{\tabcolsep}{5pt} %
\begin{tabular}{@{}llrrrrrr@{}} 
\toprule
\textbf{Model} & \textbf{Metric} & ECGFiveDay & HeartBeat & \makecell{Chlorine\\Conc.} & ECG5000 & LSST & Sleep \\
\midrule
\multirow{2}{*}{\textbf{Base Model}} 
& Params      & 3\,363      & 4\,323      & 12\,900     & 12\,966     & 29\,343     & 12\,966 \\
& F1 Score    & 0.7668      & 0.7250      & 0.4436      & 0.8886      & 0.5928      & 0.4911 \\
\midrule
\multirow{2}{*}{\textbf{Base Model$^\star$}} 
& Params      & 1\,779      & 2\,739      & 6\,660      & 6\,726      & 15\,375     & 6\,726 \\
& F1 Score    & 0.7624      & 0.7214      & 0.4440      & 0.8878      & 0.5907      & 0.4897 \\
\midrule
\multirow{2}{*}{\textbf{NdLinear}} 
& Params      & 1\,804      & 2\,752      & 6\,709      & 6\,823      & 15\,472     & 6\,823 \\
& F1 Score    & \textbf{0.7783} & \textbf{0.7363} & \textbf{0.4571} & \textbf{0.9058} & \textbf{0.6486} & \textbf{0.4978} \\
\bottomrule
\end{tabular}
\end{table*}

\begin{table}[t!]
\centering
\caption{\small \textbf{NdLinear substantially improves time-series forecasting accuracy and dramatically reduces parameter counts in both RNN and Transformer models on Electricity Transformer Temperature (ETT) datasets.} With NdLinear, model parameters are reduced by 50\%, alongside consistent gains in  accuracy. Metrics reported are mean $\pm$ standard deviation over three experimental runs.
}
\label{tab:ts_forecast_combined_side_by_side}
\resizebox{\columnwidth}{!}{%
\begin{tabular}{@{}llcccc@{}} 
\toprule
& & \multicolumn{2}{c}{RNN} & \multicolumn{2}{c}{Transformer} \\
\cmidrule(lr){3-4} \cmidrule(lr){5-6}
Dataset & Method (Params: RNN / Trans.) & MSE & MAE & MSE & MAE \\
\midrule
\multirow{2}{*}{ETTh1} & Linear (20.5k / 138k)   & 0.2900 $\pm$ 0.0170 & 0.4060 $\pm$ 0.0246 & 0.0217 $\pm$ 0.0001 & 0.1158 $\pm$ 0.0004 \\
                       & NdLinear (9.6k / 70k)   & \textbf{0.0880} $\pm$ 0.0115 & \textbf{0.2204} $\pm$ 0.0105 & \textbf{0.0173} $\pm$ 0.0003 & \textbf{0.0995} $\pm$ 0.0005 \\
\midrule
\multirow{2}{*}{ETTh2} & Linear (20.5k / 138k)   & 0.2636 $\pm$ 0.0949 & 0.3955 $\pm$ 0.0748 & 0.0226 $\pm$ 0.0001 & 0.1229 $\pm$ 0.0003 \\
                       & NdLinear (9.6k / 70k)   & \textbf{0.1536} $\pm$ 0.0137 & \textbf{0.2831} $\pm$ 0.0119 & \textbf{0.0158} $\pm$ 0.0019 & \textbf{0.0995} $\pm$ 0.0071 \\
\midrule
\multirow{2}{*}{ETTm1} & Linear (20.5k / 138k)   & 0.0187 $\pm$ 0.0012 & 0.0926 $\pm$ 0.0039 & 0.0180 $\pm$ 0.0001 & 0.1005 $\pm$ 0.0005 \\
                       & NdLinear (9.6k / 70k)   & \textbf{0.0174} $\pm$ 0.0017 & \textbf{0.0894} $\pm$ 0.0039 & \textbf{0.0161} $\pm$ 0.0006 & \textbf{0.0936} $\pm$ 0.0027 \\
\midrule
\multirow{2}{*}{ETTm2} & Linear (20.5k / 138k)   & 0.0148 $\pm$ 0.0007 & 0.0825 $\pm$ 0.0047 & 0.0151 $\pm$ 0.0000 & 0.0965 $\pm$ 0.0001 \\
                       & NdLinear (9.6k / 70k)   & \textbf{0.0139} $\pm$ 0.0009 & \textbf{0.0797} $\pm$ 0.0039 & \textbf{0.0141} $\pm$ 0.0001 & \textbf{0.0929} $\pm$ 0.0004 \\
\bottomrule
\end{tabular}%
}

\end{table}

\textbf{Results.} Table~\ref{tab:ts_tf_classify_final} in Appendix~\ref{appen:results} shows NdLinear reduces parameters by up to 47\% while 
consistently improving F1 scores across all datasets. For example, on LSST, F1 improves from 0.5928 to 0.6486 with 47\% fewer parameters. Even compared to parameter-matched baselines 
(Base Model*), NdLinear variants achieve superior performance, highlighting that the 
architectural advantage extends beyond mere parameter reduction.

\subsection{Tabular Data}
\label{sec:tabular_apps}

We evaluated NdLinear in MLPs on tabular classification (Cardio Disease dataset) \citep{cardiodisease2025} and 
regression (Food Delivery Time dataset) \cite{fooddeliverytime2025}. We replaced vanilla linear layers in two-layer 
feature extractors with NdLinear layers.

\textbf{Results.} Table~\ref{tab:ts_tf_classify_final_revised} in shows NdLinear improves performance while 
dramatically reducing parameters. For Cardio Disease classification, NdLinear achieves 
higher accuracy (73.21\% vs 72.65\%) with 67\% fewer parameters. For Food Delivery Time 
regression, NdLinear reduces MSE from 70.51 to 67.82 with 58\% fewer parameters. These 
results demonstrate that NdLinear's dimensional factorization aligns well with tabular  data where features often have independent effects.

\subsection{Computer Vision}
\label{sec:vision_apps}

\subsubsection{Vision Transformer Distillation}

We created student ViTs (NdViT) by replacing all linear layers in feedforward blocks with 
NdLinear, distilling from a pretrained ViT-B/16 teacher \citep{dosovitskiy2021image}. We tested various embedding 
dimensions (200, 300, 400) and depths (3, 6, 9 blocks) on CIFAR-10 and CIFAR-100 \citep{krizhevsky2009learning}.

\textbf{Results.} Table~\ref{tab:vit-scaling} and Figure~\ref{fig:new_combined_vit_size} in Appendix~\ref{appen:results} show NdViT consistently outperforms standard ViT 
students while using 26-68\% fewer parameters. Notably, NdViT with embedding dimension 200  surpasses a standard ViT with dimension 500, demonstrating superior parameter efficiency. 
The performance gap increases with model depth, suggesting NdLinear's benefits compound  in deeper architectures.

\subsubsection{CNN Image Classification}
\label{sec:cnn_image_class}
We replaced the penultimate linear layer in CNNs with NdLinear on CIFAR-10 and CIFAR-100, 
and compared against Tensor Regression Layers (TRL/TCL) and Tensor-Train (TT) layers.

\textbf{Results.} NdLinear achieves higher accuracy with significantly fewer parameters on both datasets (Table~\ref{tab:cnn-image-classify} in Appendix~\ref{appen:results}). On CIFAR-10, NdLinear improves accuracy by 2.6\% while reducing parameters by 94\%. On CIFAR-100, it improves accuracy by 5.1\% with 60\% fewer parameters.

\textbf{Comparison with Structured Methods.} Table~\ref{tab:trl_tcl} compares NdLinear  against TRL/TCL and TT layers on CIFAR-100. NdLinear achieves superior accuracy (71.3\%) 
compared to TRL/TCL (69.4\%) and TT (56.2\%), while using fewer parameters (434K vs 548K 
vs 769K), lower FLOPs (0.84G vs 3.97G vs 5.25G), 6× lower latency than TT (0.98ms vs 
5.87ms), and minimal GPU memory overhead (35.16 MB vs 35.60 MB vs 100.44 MB). This 
demonstrates NdLinear's practical advantages over alternative structured approaches.

\begin{table}[h]
\centering
\caption{\textbf{NdLinear outperforms existing structured tensor methods.} Comparison with Tensor Regression Layers (TRL/TCL) and Tensor-Train decomposition on CIFAR-100 classification, replacing the CNN's penultimate linear layer.}
\begin{tabular}{lccccc}
\toprule
\textbf{Method} & \textbf{Mem (MB)} & \textbf{Acc@5} & \textbf{Latency (s)} & \textbf{FLOPs (G)} & \textbf{Params} \\
\midrule
\textbf{NdLinear} & \textbf{35.16} & \textbf{0.7133} & \textbf{0.000976} & \textbf{0.843} & \textbf{433,588} \\
TRL/TCL           & 35.60          & 0.6935          & 0.001116          & 3.97            & 548,032 \\
TT                & 100.44         & 0.5617          & 0.005871          & 5.25            & 769,316 \\
\bottomrule
\end{tabular}
\label{tab:trl_tcl}

\end{table}

\subsubsection{Generative Modeling with Diffusion Transformers (DiT)}

We evaluated NdLinear in DiTs through two approaches: training from scratch on ImageNet-100~\cite{imagenet15russakovsky} and modifying pre-trained DiT models from~\cite{jin2024fast}. For training from scratch, we replaced linear layers in the 
timestep embedder. For pretrained models, we replaced linear layers in both timestep embedders and attention MLPs.

\textbf{Results.} When trained from scratch, NdLinear variants achieve lower (better) FID 
scores with comparable parameters (Figure~\ref{fig:new_combined_efficiency_showcase} in Appendix~\ref{appen:results}). For pretrained models, NdLinear 
reduces parameters from 674M to 619M while maintaining similar FID scores.
demonstrating that parameter reduction doesn't sacrifice generation quality.

\paragraph{Modifying Linear Layers in Pre-trained DiT Models} 
To evaluate performance maintenance with parameter reduction, we replaced standard Linear layers with NdLinear in pre-trained DiTs. This was done in both the timestep embedder and the MLP components of attention heads. %

\textbf{Results.} Table~\ref{tab:fid_scores} in Appendix~\ref{appen:results} shows that NdLinear-based models (619M and 563M) achieve FID scores comparable to the larger baseline 674M model, despite significant parameter reductions. %

\subsection{Ablation Studies}
\label{sec:ablations}

We conducted extensive ablations to understand NdLinear's design choices and practical considerations (full details in Appendix~\ref{app:ablations}).

\subsubsection{Design Choices}

\textbf{Per-mode Bias Terms.} Including bias terms for each dimension transformation significantly improves performance, with benefits increasing at larger widths. On the Radius Bump task, bias  terms improve MSE by 4.5\% at width 32 and 15.2\% at width 128, with negligible parameter 
overhead (<5\%) (Table~\ref{app:bias} in the Appendix). This suggests per-mode biases help capture dimension-specific offsets. More information on the taks can be found in Appendix~\ref{app:entropy}

\textbf{Axis Ordering Robustness.} NdLinear shows remarkable robustness to transformation 
order. Permuting axes (original, reverse, random) cause only 4\% accuracy variation on 
CIFAR-100, with random ordering achieving 96\% of baseline performance (Table~\ref{app:ordering} in the Appendix). This suggests NdLinear learns relatively axis-independent features, simplifying deployment.

\textbf{Hidden dimensions} offer a direct trade-off between efficiency and expressivity. Doubling hidden size improved accuracy by 2\% while still using 40\% fewer parameters than dense layers.

\subsubsection{Practical Considerations}
\textbf{Hyperparameter Robustness.} NdLinear requires no special tuning. Across learning 
rates (0.001, 0.01), batch sizes (32, 64, 128), and hidden configurations, accuracy varies only by 11 percentage points (64-75\%) on CIAFR-10 (Table~\ref{app:hyperparam} in the Appendix). Even the worst configuration (LR=0.01, batch=32) achieves 64\% accuracy with 6× fewer parameters than dense layers, demonstrating robustness  to suboptimal settings.

\textbf{Computational Overhead.} Despite sequential processing, measured overheads are 
minimal: peak memory increases by 1.1-2.0\% (CIFAR CNN: 35.17→36.91 MB), training time 
by 0.6-1.6\% (47.2→47.8s per epoch). Inference latency remains comparable due to reduced  FLOPs offsetting sequential operations (Table~\ref{app:overhead} in the Appendix).

\textbf{Sample Efficiency.} The separability bias dramatically affects data efficiency (Section~\ref{app:sample}). On 
our synthetic tasks with $\alpha=0.1$ separability, NdLinear reaches target error with 5× fewer samples 
(2,000 vs 10,000). Conversely, at $\alpha=0.9$ entanglement, it requires 1.7× more samples 
(25,000 vs 15,000), quantifying the bias-variance trade-off. The representation analysis in Section~\ref{sec:expressivity} 
provides additional mechanistic insight into these efficiency gains.

\textbf{Architectural Comparisons.} As shown in Section~\ref{sec:cnn_image_class}, NdLinear outperforms 
alternative structured layers (TRL/TCL, TT) on all metrics, validating our design choices.

\paragraph{Key Takeaway.}
NdLinear is robust to implementation details and requires no special tuning, where standard hyperparameters work well. The critical decision is architectural: whether the task exhibits sufficient separability to benefit from NdLinear's bias.

\section{Conclusion}
\label{sec:conclusion}

We introduced NdLinear, a linear layer that operates directly on N-dimensional tensors through sequential dimension-wise transformations. This simple change, by processing tensors in their native form rather than flattening them, yields dramatic improvements across modern deep learning.

Our extensive experiments demonstrate that NdLinear reduces parameters by 50-95\% while maintaining or improving performance across all tested configurations. The results reveal striking consistency: from NdLinear-LoRA fine-tuning LLaMA-8B with 9× fewer parameters while improving reasoning accuracy, to achieving 70\% lower error in time-series forecasting with half the parameters, to outperforming alternative structured methods (TRL/TCL, TT). This universality, spanning language, vision, time-series, and tabular domains, validates dimension-wise factorization as a fundamental principle.

Theoretically, we proved NdLinear maintains expressivity through preserved VC-dimension scaling, explaining why dramatic parameter reduction doesn't sacrifice performance. Our controlled experiments precisely quantified when the method's inductive bias helps versus hurts: NdLinear excels when data exhibits axis-separable structure, but struggles with highly entangled patterns. This theoretical understanding, combined with our ablations showing robustness to hyperparameters and negligible computational overhead (<2\%), provides clear deployment guidance.

Beyond its immediate practical benefits, NdLinear challenges a fundamental assumption in neural architecture design: that flattening is necessary for linear transformations. By demonstrating that structure-preserving operations consistently outperform their flattened counterparts, we reveal that the ubiquitous practice of tensor flattening has been systematically discarding valuable inductive biases. As neural networks evolve to process increasingly complex multidimensional data, from volumetric medical imaging to spatiotemporal climate models, NdLinear's principle of native dimensional processing offers not just an optimization, but a paradigm shift toward architectures that inherently respect and leverage the structure of our multidimensional world.

\clearpage
\bibliographystyle{plainnat}

\bibliography{refs}

\clearpage
\appendix

\section*{Impact Statement} 
NdLinear replaces each dense linear layer with an $N$-mode rank-1 factorization, cutting parameters and FLOPs by up to two orders of magnitude on the benchmarks evaluated in this paper. The resulting drop in compute and memory lowers energy use and enables on-device inference and federated fine-tuning, broadening access to large-model capabilities for researchers and organizations with limited hardware. The authors will release full reference code to facilitate reproducible adoption. Because cheaper deployment also reduces the barrier for misuse, future work should evaluate privacy, robustness, and bias in models that adopt this compressed design.

\section*{Limitations and Future Directions}
While NdLinear shows strong empirical efficiency, several promising research paths remain: (i) relax the rank-1 Tucker structure by permitting higher multilinear ranks or adding cross-mode residual connections to capture richer inter-mode interactions; (ii) develop memory-aware kernels that preserve compute and bandwidth efficiency as the number of modes $N$ or hidden size grows; (iii) learn or adapt the ordering of mode transforms to exploit data-driven structure; (iv) benchmark the layer in larger models ($>10\text{B}$ parameters), new modalities such as 3-D medical imaging, streaming time-series, and edge devices, measuring accuracy, latency, and memory trade-offs; (v) derive approximation-error, sample-complexity, and optimization guarantees; and (vi) investigate privacy, robustness, and fairness when NdLinear enables lightweight federated deployment. Progress on these fronts will broaden NdLinear’s applicability and deepen our understanding.

\section*{LLM Usage Disclosure}
We used large language models (LLMs) to aid and polish writing, such as improving clarity, grammar, and conciseness. We also used LLMs for retrieval and discovery, for example exhausting literature to identify potential missing related work. All technical content, proofs, experiments, and results are original contributions by the authors.

\appendix
\appendix
\section{Detailed Comparison with Related Methods}
\label{app:related2}

NdLinear's design as a structure-preserving, parameter-efficient linear layer for N-D tensors is informed by, yet distinct from, several established concepts in machine learning and tensor algebra. We detail these relationships below.

\subsection{Tensor Decomposition Methods}

NdLinear's core mechanism employs mode-wise tensor-matrix products prominent in Tucker decomposition \citep{tucker1966some, kolda2009tensor}. However, unlike classical tensor decomposition methods that primarily analyze or compress static data or pre-existing weight tensors \citep{novikov2015tensorizing, newman2024stable}, NdLinear integrates these mode-wise operations as a learnable, dynamic layer within a network. Its purpose is efficient forward transformation of activations while preserving N-D structure, not data analysis or post-hoc model compression.

NdLinear can be viewed as a \emph{hand-crafted factorization} of a fully-connected weight matrix. Specifically, the full weight matrix $W_{\text{full}}$ implicitly has a Kronecker product structure derived from mode-wise matrices $\{W_1,\ldots,W_n\}$. This corresponds to a rank-1 Tucker decomposition without a core tensor (or equivalently, a core of rank 1 in each mode).

The main trade-off is expressiveness vs.\ efficiency. NdLinear's decomposition is low-rank in a multilinear sense---it cannot represent arbitrary non-factorizable interactions between dimensions. More flexible decompositions (full Tucker or higher-rank tensor decompositions) capture more interactions but require significantly more parameters compared to NdLinear's simple sum $\sum_i D_i H_i$ and can be more challenging to train, sometimes needing special initialization or multi-stage training.

If needed, one can extend NdLinear by increasing the factorization rank (e.g., learning multiple $W_i$ matrices per mode and summing their effects, analogous to a rank-$R$ core). However, our experiments show the simple rank-1 version already performs well. Each $W_i$ clearly indicates how dimension $i$ is transformed, providing better interpretability than general tensor decomposition methods that disperse transformations across multiple factors.

\subsection{Factorized Convolutions and Axial Operations}

Neural nets have long exploited \emph{axis-wise} structure to cut parameters and impose useful priors. In CNNs, an image tensor $X\!\in\!\mathbb{R}^{H\times W\times C}$ is \emph{not} flattened; instead, convolutions slide local kernels over the two spatial axes $(H,W)$ while mixing channels $C$. In sequence models, a tensor $X\!\in\!\mathbb{R}^{L\times C}$ (length $L$, features $C$) is processed along the \emph{sequence axis} $L$ (e.g., attention/conv) while mixing feature \emph{channels} $C$ within positions. Below we recap common factorized operators and their limitations.

\paragraph{Definitions.}
\emph{Channels} are feature maps carried alongside a position-like axis (e.g., RGB planes or intermediate filters in CNNs; hidden features per token in sequences). \emph{Sequence (or spatial) dimensions} are position-like axes along which locality or ordering matters (e.g., time $L$, image height $H$, width $W$).

\paragraph{Grouped convolutions.}
Grouped convs restrict each filter to operate on a subset of channels, partitioning $C$ into groups to reduce parameters and FLOPs \citep{krizhevsky2012imagenet, xie2017aggregated}.
\emph{Limitation:} channel mixing is constrained within groups; cross-group interactions require additional layers, and the grouping choice is a manual architectural prior.

\paragraph{Depthwise separable convolutions.}
Depthwise separable convs factor a full conv into a \emph{depthwise} spatial conv per channel followed by a $1{\times}1$ (\emph{pointwise}) conv that mixes channels \citep{chollet2017xception, howard2017mobilenets}. This yields large FLOP/parameter savings with strong accuracy.
\emph{Limitations:} the factorization is tied to 2D spatial structure; expressivity hinges on the pointwise mixer; extending beyond standard spatial axes typically needs custom kernels.

\paragraph{Axial (factorized) attention.}
Axial attention decomposes 2D/3D attention into a sequence of 1D attentions applied along one axis at a time (e.g., height then width), cutting quadratic costs while preserving long-range interactions along each axis \citep{ho2019axial, wang2020axial, yan2023tt}.
\emph{Limitations:} axis order becomes an architectural prior; full joint interactions across axes emerge only after stacking, and costs can still be high for very long axes.

\paragraph{CNNs.}
CNNs are the canonical instance of \emph{axis-aware} processing: they exploit spatial locality (factorization in $(H,W)$) and defer heavy channel mixing to $1{\times}1$ pointwise layers. Grouped/depthwise variants intensify this factorization to further reduce compute.

\paragraph{NdLinear in this landscape.}
NdLinear generalizes the factorization principle beyond conv/attention mechanics: given an $n$-D activation $X\!\in\!\mathbb{R}^{D_1\times\cdots\times D_n}$, it \emph{systematically factorizes a linear map} into mode-wise transforms $D_i\!\to\!H_i$ and applies them \emph{sequentially}, producing a structured output $Y\!\in\!\mathbb{R}^{H_1\times\cdots\times H_n}$. Thus, instead of a single dense matrix $\mathbb{R}^{\prod_i D_i}\!\to\!\mathbb{R}^{\prod_i H_i}$, NdLinear uses $n$ small matrices with $\sum_i D_i H_i$ parameters, preserving the N-D shape throughout. Unlike grouped/depthwise convs (tied to spatial kernels) or axial attention (tied to attention mechanics and axis ordering), NdLinear:
(i) applies to \emph{arbitrary} N-D tensors (images, videos, spectrograms, multivariate time series, tabular tensors), 
(ii) naturally supports both \emph{compression and expansion} per mode $(H_i \lessgtr D_i)$,
(iii) creates insertion points for normalization/activation between mode maps, and
(iv) preserves a clear axis-wise inductive bias without requiring handcrafted groups or convolutional kernels.

\paragraph{Known failure modes and how NdLinear relates.}
Axis-factorized operators can underperform when tasks require strong \emph{entangled} cross-axis interactions (e.g., patterns tied jointly to $(H,W)$ rather than separably to each). Grouped/depthwise convs may also struggle if the chosen grouping misaligns with semantics; axial attention can be sensitive to axis order and depth. NdLinear shares the core trade-off (axis-wise separability vs.\ full expressivity) but makes it \emph{explicit and tunable} by per-mode widths/ranks or stacking; it inherits the efficiency benefits of factorization while remaining N-D agnostic in form.

\subsection{Tensor Contraction and Regression Layers}

\paragraph{Tensor Contraction Layers (TCL).}
TCLs \citep{cichocki2014era, novikov2015tensorizing} implement a \emph{one-shot} multilinear map by contracting an input tensor $X\!\in\!\mathbb{R}^{D_1\times\cdots\times D_m}$ with a set of mode matrices $\{V^{(k)}\in\mathbb{R}^{R_k\times D_k}\}_{k=1}^m$, yielding a reduced tensor
$X \times_1 V^{(1)} \times_2 \cdots \times_m V^{(m)} \in \mathbb{R}^{R_1\times\cdots\times R_m}$.
The goal is typically \emph{dimensionality reduction} (feature compression) via tensor contractions prior to downstream layers; intermediate positions for normalization/activation and per-mode biases are usually not part of the basic formulation \citep{cichocki2014era, novikov2015tensorizing}.

\paragraph{Tensor Regression Layers (TRL).}
TRLs \citep{kossaifi2020tensor} cast prediction as supervised regression/classification with a \emph{fixed tensor format} for the weight (e.g., Tucker-/TT-structured). Given an input tensor $X$, a TRL fits a low-rank tensor $W$ (plus optional bias) such that $\langle W, X\rangle$ (or a nonlinear variant) matches targets. The emphasis is on \emph{learning with low-rank weights} for sample/parameter efficiency; the tensor format (ranks/cores) is chosen a priori and does not expose interleaved normalization/activation between modes \citep{kossaifi2020tensor}.

\paragraph{NdLinear vs.\ TCL/TRL.}
NdLinear is a \emph{structure-preserving, learnable linear layer} that applies \emph{sequential} mode-wise maps to activations,
\(
Y \;=\; X \times_1 W_1 \times_2 \cdots \times_m W_m \in \mathbb{R}^{H_1\times\cdots\times H_m},
\)
with an implicit rank-1 Tucker/Kronecker weight and $\sum_k D_k H_k$ parameters. This design differs in intent and mechanics:
\begin{itemize}[topsep=2pt,itemsep=1pt,parsep=0pt]
\item \textbf{Goal.} TCL primarily \emph{contracts} modes for reduction \citep{cichocki2014era, novikov2015tensorizing}; TRL fixes a low-rank \emph{regression} format \citep{kossaifi2020tensor}; NdLinear is a \emph{drop-in linear} alternative for N-D activations that preserves the full tensor shape and supports \emph{expansion or compression} per mode $(H_k \gtrless D_k)$.
\item \textbf{Execution.} TCL performs a \emph{single} contraction; NdLinear performs \emph{sequential} per-mode maps, creating natural insertion points for LayerNorm/Dropout/activations and allowing \emph{per-mode biases}. TRL optimizes a fixed low-rank weight but does not expose interleaved, mode-by-mode transforms during the forward pass.
\item \textbf{Expressivity/control.} All three impose structured priors; NdLinear’s axis-wise separability can be \emph{tuned} via widths/ranks or stacked blocks, interpolating between strong separability and near-dense behavior while retaining N-D outputs.
\end{itemize}

\paragraph{TCL as a special case of NdLinear.}
Under the joint constraints \emph{(i)} $H_k{=}R_k$ for all $k$ (no expansion beyond the contracted size), \emph{(ii)} no per-mode biases, and \emph{(iii)} no interleaved operations between mode maps (i.e., a single, commutative product), the one-shot TCL contraction coincides with NdLinear \citep{cichocki2014era, novikov2015tensorizing}. Outside this narrow corner, NdLinear is strictly more general and practical: it preserves N-D structure through sequential maps, supports per-mode biases and interleaving (stability), and flexibly expands or compresses each mode.

\subsection{Structured Linear Layers}

NdLinear sits within the broader family of \emph{structured} linear layers for parameter efficiency \citep{denil2013predicting, wei2024building}. Classic approaches constrain a large dense weight $W\!\in\!\mathbb{R}^{(\prod_i D_i)\times(\prod_i H_i)}$ by imposing algebraic structure \emph{on $W$ itself}—e.g., low-rank factorizations \citep{sainath2013low}, block/Butterfly/Monarch-style sparse–fast transforms \citep{dao2022monarch}, or other learned structured matrices \citep{sindhwani2015structured, potapczynski2024searching}. These methods decouple the parameterization of $W$ from the native organization of the activations, often yielding strong compression but requiring the model to implicitly discover how that structure aligns with the data.

In contrast, NdLinear derives its structure \emph{from the N-D layout of the input}. Given $X\!\in\!\mathbb{R}^{D_1\times\cdots\times D_n}$, NdLinear replaces the dense map $\mathbb{R}^{\prod_i D_i}\!\to\!\mathbb{R}^{\prod_i H_i}$ with \emph{sequential mode-wise transforms} $D_i\!\to\!H_i$, preserving tensor shape and inducing an explicit Kronecker (rank-1 Tucker) weight with only $\sum_i D_i H_i$ parameters. This data-centric factorization (i) makes the axis-wise inductive bias transparent and tunable (via per-mode widths/ranks or stacking), (ii) supports per-mode expansion or compression $(H_i \gtrless D_i)$, and (iii) exposes natural insertion points for normalization/activation between modes—while retaining the efficiency benefits typical of structured linear layers \citep{denil2013predicting, wei2024building, sainath2013low, dao2022monarch, sindhwani2015structured, potapczynski2024searching}.

\subsection{Graph Neural Networks}

Graph Neural Networks (GNNs) address data with \emph{irregular} connectivity by propagating information over edges via message passing \citep{scarselli2008graph, micheli2009neural, bronstein2017geometric, zhou2020graph, wu2020comprehensive}. Concretely, a GNN updates node features by aggregating messages from neighbors and applying learnable transforms \citep{gilmer2017neural, kipf2017semi, hamilton2017inductive}. Stacking layers increases the receptive field and enables global interaction, but typically requires multiple rounds of propagation to mix distant nodes \citep{battaglia2018relational}.

While an N-D grid (e.g., image or tensor lattice) \emph{can} be modeled as a graph (one node per cell, edges to local neighbors), this introduces unnecessary overhead on regular grids: message passing is inherently local, so achieving global mixing along each axis often demands many layers, with added memory/compute and potential optimization issues (e.g., depth-related bottlenecks). Moreover, parameter sharing in GNNs is tied to edge types and neighborhood schemas, not directly to axis-wise tensor structure.

NdLinear takes the complementary route for \emph{regular tensor grids}. Given $X\!\in\!\mathbb{R}^{D_1\times\cdots\times D_n}$, it applies \emph{global, mode-wise linear maps} $D_i\!\to\!H_i$ in a single layer, mixing information \emph{along entire axes} without constructing a graph or iterating local messages. This preserves the N-D layout, yields $\sum_i D_i H_i$ parameters via an implicit Kronecker (rank-1 Tucker) structure, and exposes insertion points for normalization/activation between mode maps. In short: GNNs excel when connectivity is irregular or non-Euclidean \citep{bronstein2017geometric, zhou2020graph}, whereas NdLinear specializes in axis-aware, structure-preserving transformations on regular tensors, providing simpler and often faster global mixing along each dimension.

\paragraph{When to use which.}
Use GNNs for arbitrary graphs, heterogeneous edge types, and relational reasoning on non-grid data \citep{scarselli2008graph, micheli2009neural, bronstein2017geometric, zhou2020graph, wu2020comprehensive}. Use NdLinear when data are naturally arranged as regular tensors and you want efficient, axis-wise global interactions without flattening; its bias toward axis separability provides parameter savings and predictable behavior on grid-structured domains.

\subsection{Other Specialized Approaches}

A variety of specialized architectures preserve structure without resorting to full flattening:

\paragraph{Slicing-based layers.}
Methods that slice inputs along spatial/temporal (or rotated) subdomains process each slice with shared weights, then recombine \citep{shao2016slicing, dieleman2016exploiting}. This preserves locality and orientation information with modest compute.
\emph{Limitations:} boundaries between slices can hinder cross-slice interaction; designs are task-/geometry-specific and often require bespoke preprocessing.

\paragraph{Capsule Networks.}
Capsules use vector/matrix-valued units and routing to model part–whole hierarchies and pose relationships \citep{hinton2011transforming, sabour2017dynamic, hinton2018matrix}. They maintain structured representations through learned agreement between capsules.
\emph{Limitations:} routing adds iterative, nontrivial overhead; scaling to large resolutions and datasets has proven challenging; design choices (routing, capsule size) are sensitive.

\paragraph{Hadamard/Fourier feature mixing.}
Fixed orthogonal or Fourier-like transforms provide global mixing with $\mathcal{O}(N\log N)$ or even $\mathcal{O}(N)$ cost (e.g., Random Features, Fastfood, FNet, Block-based variants) \citep{rahimi2007random, le2013fastfood, tancik2020fourier, lee2022fnet, pan2022block}.
\emph{Limitations:} transforms are \emph{fixed} (non-learnable) or only weakly parameterized, so alignment with data structure must be recovered by subsequent layers; expressivity depends on depth.

\paragraph{Relation to NdLinear.}
NdLinear applies \emph{learned}, factorized linear maps along each tensor mode, preserving the full N-D layout while enabling axis-wise global mixing with $\sum_i D_i H_i$ parameters. Unlike slicing \citep{shao2016slicing, dieleman2016exploiting}, it does not require hand-crafted partitions; unlike capsules \citep{hinton2011transforming, sabour2017dynamic, hinton2018matrix}, it avoids iterative routing; unlike fixed Hadamard/Fourier mixers \citep{rahimi2007random, le2013fastfood, tancik2020fourier, lee2022fnet, pan2022block}, it learns mode-wise transforms end-to-end. This yields a simple, geometry-agnostic mechanism for structure-preserving, parameter-efficient linear transformation on regular tensors.

\section{More Related Work}
\label{sec:related_work}

Modern neural networks contain substantial parameter redundancy: a large fraction of weights can be predicted from a small subset, sometimes up to 95\% with no loss in accuracy \citep{denil2013predicting}. This has motivated a broad line of work on \emph{efficient parameterizations} that preserve accuracy while reducing storage and compute.

\paragraph{Structured tensor factorization.}
A major thread leverages high-order structure via tensor decompositions of weights. CP/Tucker-style compressions applied to convolutional kernels reduce parameters and inference cost \citep{Lebedev2015}. Tensor Train (TT) layers compress fully-connected mappings into compact tensorized operators \citep{novikov2015tensorizing}. Block-Term (BT) tensor networks combine Tucker- and CP-like structure for additional flexibility \citep{Ye2020}. These tensor-structured layers reduce parameters while retaining rich multi-way interactions by factoring weights across modes.

\paragraph{Structured matrices and parameter sharing.}
Another approach imposes algebraic structure on large dense matrices, replacing them with families that admit fast transforms and fewer degrees of freedom. Toeplitz-like and related structured operators provide strong compression with competitive accuracy \citep{sindhwani2015structured}; related families (e.g., circulant, block-circulant) and low-rank factorizations likewise trade unrestricted expressivity for parameter/compute efficiency \citep{Lebedev2015}.

\paragraph{Multi-space representations.}
Complementary to structural compression, multi-space learning embeds features into multiple geometries to better capture hierarchy and long-range relations. For example, jointly using Euclidean and hyperbolic spaces for LiDAR yields improved hierarchical encoding and pose estimation \citep{Wang2023}. Such representations enhance expressivity without necessarily increasing individual layer sizes.

\paragraph{Preserving high-order structure in practice.}
Operational layers that respect native tensor axes often strike favorable accuracy–efficiency trade-offs. Depthwise separable convolutions split channel-wise and spatial mixing to cut FLOPs while preserving inductive bias \citep{chollet2017xception, howard2017mobilenets}. However, many fully-connected stages still flatten activations, discarding axis structure learned upstream.

\paragraph{Positioning NdLinear.}
NdLinear aligns with these trends but differs in where structure is imposed: rather than factorizing \emph{weights after flattening}, it performs \emph{mode-wise} learned linear maps directly on N-D activations, preserving tensor layout throughout. Conceptually, it is a rank-1 Tucker (Kronecker) parameterization of the dense linear map, with parameters scaling as $\sum_i D_i H_i$ rather than $\prod_i D_i \prod_i H_i$. This data-aligned factorization complements tensorized weights \citep{Lebedev2015, novikov2015tensorizing, Ye2020} and structured matrices \citep{sindhwani2015structured}, and, like depthwise separable convolutions \citep{chollet2017xception, howard2017mobilenets}, leverages axis-aware inductive bias—without resorting to flattening.

\section{Proofs and Technical Details}
\label{app:proofs}

\subsection{VC-Dimension Analysis}
\label{app:vc}

We analyze the expressive capacity of NdLinear compared to standard linear layers. Following \citet{bartlett2019nearly}, any piecewise-linear feedforward network with $P$ parameters has VC-dimension $\Theta(P \log P)$.

\begin{theorem}[VC-Dimension of NdLinear]
\label{thm:vcdim}
Consider input tensors of shape $(B, a, b, c)$ transformed to outputs of shape $(B, d, d, d)$. Let:
\begin{itemize}
    \item $N_{\text{vanilla}} = dabc$ (parameters in vanilla linear layer)
    \item $N_{\text{nd}} = d(a + b + c)$ (parameters in NdLinear)
\end{itemize}
Then:
\begin{enumerate}
    \item As $d \to \infty$ with $a, b, c$ fixed: $N_{\text{nd}} = \Theta(N_{\text{vanilla}})$
    \item NdLinear's VC-dimension is $\Theta(N_{\text{nd}} \log N_{\text{nd}})$
\end{enumerate}
\end{theorem}

\begin{proof}
For part (1), observe that:
\begin{align*}
N_{\text{nd}} = d(a + b + c) = \frac{a + b + c}{abc} \cdot dabc = \frac{a + b + c}{abc} \cdot N_{\text{vanilla}}.
\end{align*}

Since $\frac{a + b + c}{abc}$ is a positive constant (for fixed $a, b, c$), we have:
\begin{align*}
N_{\text{nd}} = \Theta(N_{\text{vanilla}}) \quad \text{as } d \to \infty.
\end{align*}

For part (2), by the Bartlett et al. result, since NdLinear has $N_{\text{nd}}$ parameters and maintains piecewise-linear structure through ReLU activations:
\begin{align*}
\text{VCdim}_{\text{NdLinear}} = \Theta(N_{\text{nd}} \log N_{\text{nd}}).
\end{align*}
\end{proof}

\textbf{Interpretation:} While NdLinear uses fewer parameters for finite $d$, as the hidden dimension grows, its parameter count becomes proportional to vanilla linear layers, preserving the same VC-dimension scaling.

\begin{theorem}[Parameter Count Lower Bound]
\label{thm:lowerbound}
For positive integers $a, b, c, d$:
\begin{align*}
d(a + b + c) > \log(dabc).
\end{align*}
\end{theorem}

\begin{proof}
We have:
\begin{align*}
\log(dabc) &= \log d + \log a + \log b + \log c \\
&\leq (d - 1) + (a - 1) + (b - 1) + (c - 1) \quad \text{(using } \log x \leq x - 1\text{)} \\
&= d + a + b + c - 4.
\end{align*}

Therefore:
\begin{align*}
d(a + b + c) - \log(dabc) &\geq d(a + b + c) - (d + a + b + c - 4) \\
&= (d - 1)(a + b + c) - d + 1 + 4 \\
&= (d - 1)(a + b + c - 1) + 4 \\
&\geq 4 > 0,
\end{align*}
since $d \geq 1$ and $a + b + c \geq 3$ for non-trivial tensors.
\end{proof}

\subsection{Peak Activation Memory Analysis}
\label{app:memory}

\begin{proposition}[Peak Memory Overhead Bound]
For an input tensor with $m$ modes of sizes $(d_1, \ldots, d_m)$ where $\prod_{i=1}^m d_i = D$, and output sizes $k_i \leq d_i$, the peak additional activation memory for backpropagation satisfies:
\begin{align*}
\frac{\text{extra activation memory}}{\text{baseline activation memory}} \leq \frac{\max_i (k_i/d_i) \cdot \min_i d_i}{m \cdot \min_i d_i} \leq \frac{1}{m}.
\end{align*}
For typical 3D tensors (m=3), this overhead is at most 33\%.
\end{proposition}

\begin{proof}
During forward pass, NdLinear sequentially transforms each mode. For backpropagation, we store intermediate activations after each mode transformation.

After transforming $j$ modes, the tensor has shape:
\begin{align*}
(B, k_1, \ldots, k_j, d_{j+1}, \ldots, d_m).
\end{align*}

The peak extra memory occurs at the stage with the largest intermediate tensor. Since $k_i \leq d_i$, each intermediate tensor has at most $BD$ elements. The baseline memory is also $BD$ elements.

In the worst case where all $k_i = d_i$, we have at most $(m-1)$ intermediate tensors to store, but only one is needed at any given time during backprop (due to sequential processing). Therefore:
\begin{align*}
\text{Peak overhead} = \frac{BD \cdot \max_i(k_i/d_i)}{BD} \leq \frac{1}{m}.
\end{align*}
\end{proof}

\subsection{Computational Complexity}
\label{app:flops}

\begin{proposition}[Exact FLOP Count]
NdLinear transforming $\mathcal{X} \in \mathbb{R}^{B \times D_1 \times \cdots \times D_N}$ to $\mathcal{Y} \in \mathbb{R}^{B \times H_1 \times \cdots \times H_N}$ requires:
\begin{align*}
\text{FLOPs}_{\text{NdLinear}} = 2B \sum_{k=1}^{N} \left[\left(\prod_{j=1}^{k-1} H_j\right) \left(\prod_{j=k+1}^{N} D_j\right) D_k H_k\right],
\end{align*}
where the factor of 2 accounts for multiply-add operations.
\end{proposition}

This is typically $\mathcal{O}(BND^{N+1})$ for $D_i = H_i = D$, compared to $\mathcal{O}(BD^{2N})$ for vanilla linear layers—yielding orders of magnitude savings as $N$ increases.

\section{Implementation Details and Training Protocol}
\label{sec:implementation}
Here we present implementation details and training protocol of NdLinear (Algorithm \ref{alg:ndlinear}).
Implementing NdLinear in practice involves careful attention to efficiency and compatibility with existing deep learning frameworks. 
We outline key considerations in the following.

\textbf{Memory Efficiency.} 
Despite handling high-dimensional tensors, NdLinear is \textit{memory-efficient} due to its factorized parameterization. The forward pass requires allocating intermediate tensors during each mode transformation (after each linear operation, the tensor has one updated dimension). However, these intermediate allocations are of the same order as the input/output size and significantly smaller than the memory required for a gigantic flattened weight matrix. Modern tensor libraries (PyTorch, TensorFlow) facilitate implementing transpose-reshape-multiply steps without excessive data copying. We ensure in-place operations where possible (e.g., using $\mathtt{view}$ in PyTorch). Operations primarily reuse the input buffer for output as each mode is transformed, ensuring modest peak memory usage.

\textbf{Parameter Initialization.} Each weight matrix $W_i$ can be initialized using standard strategies for linear layers (Xavier/Glorot \cite{glorot2010understanding} or Kaiming \cite{he2015delving} initialization based on fan-in and fan-out). Since $W_i$ has fan-in $D_i$ and fan-out $H_i$, the initialization follows
${\rm Uni}(-\sqrt{\frac{6}{D_i+H_i}}, \sqrt{\frac{6}{D_i+H_i}})$ for Xavier uniform, or analogous formulas for other initializations. This helps maintain stable gradients across modes. One subtle point is that if $n$ is large, each mode’s weight is relatively small, mitigating the risk of extremely large fan-in. We observed no initialization-specific difficulties; indeed, NdLinear's parameter reduction may help avoid gradient explosion or vanishing issues in deep networks. 

\textbf{Computational Overhead.} 
Factorized operations (multiple transpose and reshape operations with smaller matrix multiplications) are highly optimized in modern BLAS libraries. Practically, runtime is comparable to or faster than fully-connected layers with similar outputs, due to reduced total FLOPs. Python-level overhead is minimal; the algorithm can be implemented in a single forward function looping over modes. For moderate $n$ (up to 4 or 5 dimensions), this loop is short. Explicit loops or unrolling (NdLinear2d, NdLinear3d, etc.) are feasible, but a simple loop suffices. Autograd engines handle tensor operations seamlessly, allowing standard backpropagation. Each $W_i$ receives gradients normally from upstream gradients.

\textbf{Training Protocols.}
NdLinear layers can be trained end-to-end with standard optimization algorithms (SGD, Adam) just like standard linear layers. 
Loss functions depend on the task (cross-entropy, MSE, etc.) and are unaffected by NdLinear. 
However, because NdLinear significantly reduces parameters compared to fully-connected layers, it tends to overfit less, possibly needing less aggressive regularization. 
Common techniques remain useful: \textit{weight decay} (L2 regularization) on weights $W_i$, and optionally \textit{dropout} between layers. 
Dropout can be applied before or after NdLinear; entries in the output tensor $Y$ can be dropped as usual. Specialized regularizers for factorized weights (norm regularization, orthogonality) may help further restrict solution spaces, though not required.

\textbf{Optimization and Convergence.} 
Practically, each $W_i$ is updated based on a portion of the overall error gradient (due to sequential mode transforms). 
In experiments, all $W_i$ matrices learned smoothly with default optimizer settings. 
If dimensionality varies significantly across modes, gradient clipping or adaptive learning rates per mode may be beneficial.
Throughout our numerical investigations, training dynamics are stable overall --- NdLinear layers integrated seamlessly into models without requiring special tuning. 
Standard protocols (learning rate schedules, early stopping criteria, etc.) used for equivalent models with dense layers apply here.

\section{Full Experimental Details and Results}
\label{appen:results}
We present full experimental details and more results on LoRA fine-tuning in \cref{appen:results:lora}; language-model pretraining (OPT and BERT) in \cref{appen:OPT}; time-series prediction (RNN and Transformer) in \cref{appen:time_series}; tabular data in \cref{appen:tabular}; and vision tasks (CNN, ViT, and DiT) in \cref{appen:vision}. The complete experiment code is available at \url{https://github.com/cyclone-trout/ndlinear_neurips}.

\subsection{Parameter-Efficient Finetuning with LoRA and NdLinear-LoRA}
\label{appen:results:lora}
In our study, we utilized state-of-the-art transformer architectures to investigate the impact of targeted modu le adaptation. We selected two base models, Qwen3-1.7B-Base \citep{yang2025qwen3} and Meta-Llama-3-8B \citep{dubey2024llama3}, recognized for their robust performance across various tasks. To focus our adaptations, we targeted specific modules within these models, including \texttt{q\_proj}, \texttt{k\_proj}, \texttt{v\_proj}, \texttt{o\_proj}, \texttt{gate\_proj}, \texttt{up\_proj}, and \texttt{down\_proj}. This approach allowed us to enhance model capacity and efficiency selectively.

For the adaptation process, we employed Low-Rank Adaptation (LoRA) techniques \citep{hu2021lora}, specifically using both NdLinear LoRA and classic LoRA configurations. We explored a range of alpha values (1, 4, and 8) and rank settings (1, 4, and 8) to determine the optimal configuration for our models. This exploration was critical for understanding how different levels of parameter sharing and scaling affect model performance and generalization.

The training process was conducted using the AdamW optimizer, a choice informed by its effectiveness in managing the complexities of transformer models. We set the learning rate to $1 \times 10^{-4}$, which provided a suitable balance between convergence speed and training stability. The batch size was set to 1, a decision that facilitated the use of gradient accumulation to optimize GPU memory usage. To ensure the models could handle a wide variety of inputs, we set the maximum sequence length to 512 tokens. The models were trained over 2 epochs, a duration found to be sufficient for achieving significant performance improvements without excessive computational cost. To ensure reproducibility, we used a random seed of 42 across all experiments.

Our models were fine-tuned using two datasets: Math10K and CommonsenseQA. These datasets were chosen for their ability to challenge the models with both mathematical reasoning and commonsense understanding. For evaluation, we employed a diverse set of benchmark datasets, including GSM8K, MultiArith, ARC-C, ARC-E, and BoolQ. This selection allowed us to assess the models' generalization capabilities across different types of reasoning tasks.

The entire implementation was carried out on a single NVIDIA H100 GPU, using Hugging Face’s \texttt{AutoModelForCausalLM} framework\footnote{\url{https://huggingface.co/docs/transformers/en/model_doc/auto}}, integrated with our custom \texttt{NdLinear} adapter layer. Datasets were tokenized using the default tokenizer for each model, with padding applied to the \texttt{eos\_token}. We employed label masking to exclude prompt tokens from loss computation, ensuring that training focused on the relevant portions of the input. Our implementation leveraged PyTorch, along with Hugging Face Transformers, PEFT, and Accelerate, to facilitate efficient model training and adaptation. Evaluation was performed in a zero-shot setting using greedy decoding, which provided a consistent measure of model performance without the variability introduced by sampling methods.

\label{appen:OPT}
\textbf{Open Pretrained Transformer (OPT) \citep{zhang2022opt}.} For OPT-Small, which originally contains 124M parameters, replacing the standard linear layers with NdLinear reduces the parameter count to {119M}. Similarly, for OPT-Mid, the parameter count decreases from 350M to {337M} after the substitution.

\begin{table}[!ht]
    \centering
    \caption{\small Perplexity comparison for OPT-Small and OPT-Mid models with Linear vs. NdLinear layers. }
    \begin{tabular}{lcc}
        \toprule
        & {Linear} & {NdLinear} \\
        \midrule
        {OPT-Small (Params)} & 15.970 (124M) & \textbf{15.755} (\textbf{119M}) \\
        {OPT-Mid (Params)} & 12.926 (350M) & \textbf{12.565} (\textbf{337M}) \\
        \bottomrule
    \end{tabular}
    \label{tab:linear_vs_ndlinear}
\end{table}

\begin{figure}[ht]
    \centering
    \begin{subfigure}[b]{0.44\textwidth}
        \centering
        \includegraphics[width=\textwidth]{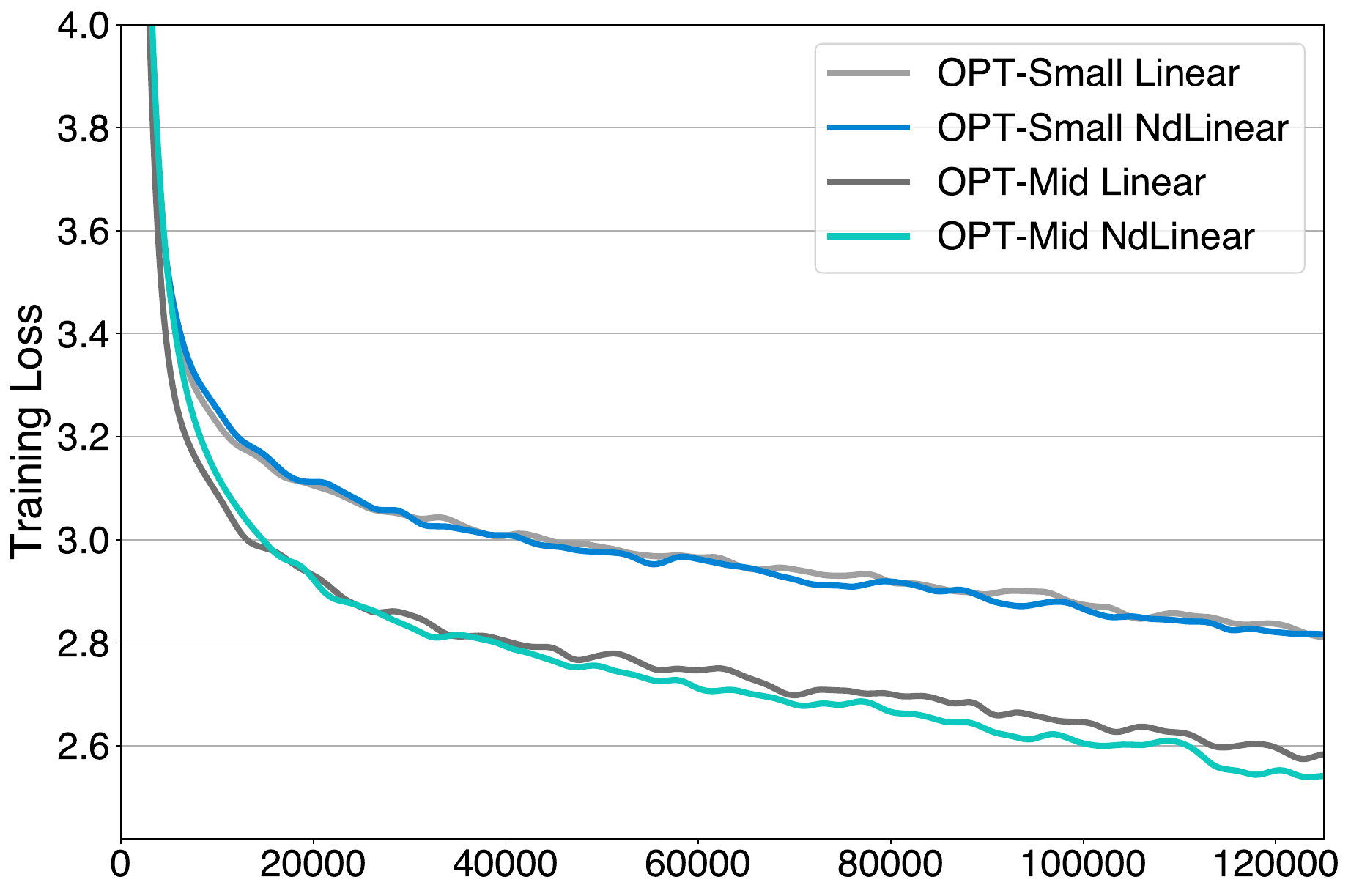}
        \caption{\small Training Loss during Pre-Training.}
        \label{fig:opt_train}
    \end{subfigure}
    \hfill %
    \begin{subfigure}[b]{0.44\textwidth}
        \centering
        \includegraphics[width=\textwidth]{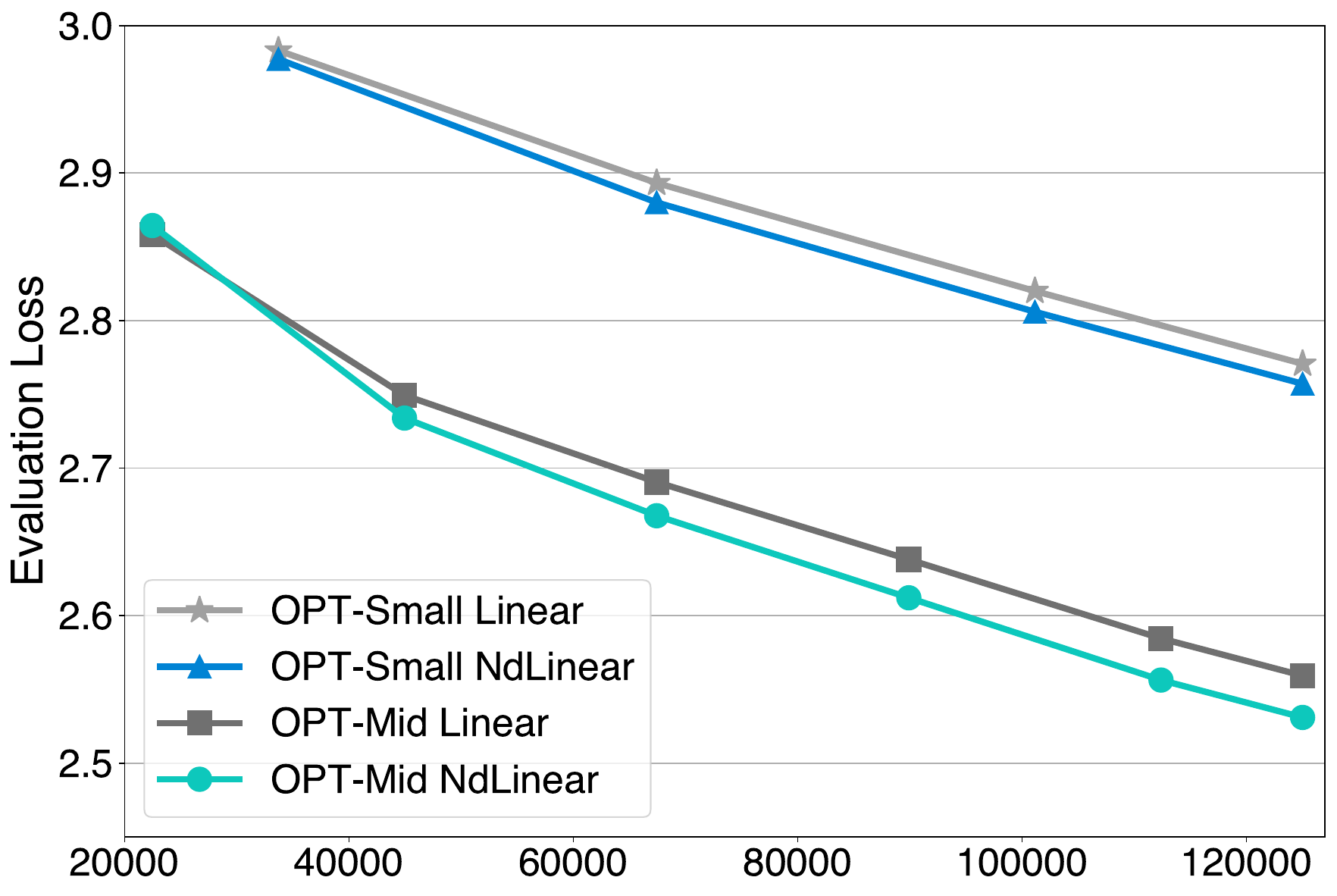} 
        \caption{\small Evaluation Loss during Pre-Training.}
        \label{fig:opt-eval}
    \end{subfigure}
    \caption{\small Training and evaluation loss curves during OPT model pretraining. NdLinear variants consistently achieve lower loss values. x-axis represents the number of training steps.}
    \vspace{-1.5em}
    \label{fig:opt-pretrain-loss-curve}
\end{figure}

In Table \ref{tab:linear_vs_ndlinear}, both the OPT-Small and OPT-Mid models achieve lower perplexity scores after replacing standard linear layers with NdLinear layers, despite having fewer parameters. Moreover, the performance improvement becomes more significant as model size increases, with the perplexity gap widening from $0.215$ in OPT-Small to $0.361$ in OPT-Mid. Figure \ref{fig:opt-pretrain-loss-curve} shows that OPT models with NdLinear feedforward layers achieve lower final training and evaluation losses compared to their counterparts using standard linear feedforward layers.

\textit{Zero-shot Tasks. } We also evaluate the OPTs' pretraining on 10 zero-shot NLP tasks:
\begin{itemize}[left=0.25cm]
    \item \textbf{Natural Language Inference Tasks}: CB \cite{de2019commitmentbank}
    \item \textbf{Coreference Resolution Tasks}: Winogrande \cite{sakaguchi2021winogrande} 
    \item \textbf{Sentence Completion Tasks}: COPA \cite{roemmele2011choice}, HellaSwag \cite{zellers2019hellaswag}
    \item \textbf{Word Sense Disambiguation Tasks}: WiC \cite{pilehvar2018wic}
    \item \textbf{Question Answering Tasks}: 
    ARC-Easy, ARC-Challenge \cite{clark2018think}, 
    OpenBookQA \cite{mihaylov2018can}, 
    BoolQ \cite{clark2019boolq}
    \item \textbf{Commonsense Reasoning Tasks}: PIQA \cite{bisk2020piqa}
\end{itemize}
During evaluation, we cast all of the above tasks into a \textit{multiple-choice format}.
Namely, the goal is to select the correct completion from a set of candidate options. 
For each option, we compute the language model (LM) likelihood of the full input consisting of the context concatenated with the candidate completion.
To account for differences in the lengths of candidate completions, we compute the average per-token log-likelihood for each option, following  \cite{brown2020language}. 
The model's prediction is taken to be the option with the highest per-token likelihood.
\begin{table}[h]
\centering
\caption{\small Perplexity Score and Zero-Shot Performance on OPT Model with and without NdLinear. }
\renewcommand{\arraystretch}{1.15}
\setlength\extrarowheight{0pt}
\begin{tabular}{ccccc}
\toprule
\multirow{2}{*}{\begin{tabular}[c]{@{}c@{}}\centering {\bf }\end{tabular}} 
& \multicolumn{2}{c|}{{\bf OPT-Small}} 
& \multicolumn{2}{c}{{\bf OPT-Mid}} \\ \cmidrule(lr){2-5} 
& \multicolumn{1}{c}{{\bf $\,\,\,\,\,$Linear$\,\,\,\,\,$}} & \multicolumn{1}{c|}{{\bf NdLinear}} 
& {\bf $\,\,\,\,\,$Linear$\,\,\,\,\,$} & {\bf NdLinear} \\ \midrule
Num of Params & 
124M & 
\textbf{119M} & 
350M & 
\textbf{337M} \\ \midrule
Perplexity & 
15.970 & 
\textbf{15.755} & 
12.926 & 
\textbf{12.565} \\ \midrule
CB &
0.32 &
\textbf{0.38} &
0.50 &
\textbf{0.52} \\ \midrule
Winogrande &
0.49 &
\textbf{0.51} &
\underline{0.50} &
\underline{0.50} \\ \midrule
COPA &
\textbf{0.58} &
0.53 &
0.54 &
\textbf{0.56} \\ \midrule
HellaSwag &
\underline{0.26} &
\underline{0.26} &
0.28 &
\textbf{0.29} \\ \midrule
WiC &
\textbf{0.51} &
0.50 &
\textbf{0.50} &
0.49 \\ \midrule
ARC-Easy &
0.29 &
\textbf{0.30} &
\underline{0.29} &
\underline{0.29} \\ \midrule
ARC-Challenge &
0.24 &
\textbf{0.25} &
\textbf{0.24} &
0.23 \\ \midrule
OpenBookQA &
0.32 &
\textbf{0.35} &
\underline{0.34} &
\underline{0.34} \\ \midrule
BoolQ &
0.49 &
\textbf{0.59} &
\textbf{0.60} &
0.56 \\ \midrule
PIQA &
\textbf{0.53} &
0.50 &
\underline{0.53} &
\underline{0.53} \\ 
\bottomrule
\end{tabular}
\label{tab:opt-pretrain}
\end{table}

\textbf{BERT \citep{devlin2019bert}.} We replace the conventional two-layer linear classification head in BERT with an NdLinear layer followed by a classification layer. The NdLinear transforms have hidden dimensions of $(2,2)$. Each model is trained for 200 epochs with a batch size of 32, a hidden layer size of 128, and a learning rate of 0.005.

\begin{table}[h]
\centering
\small
\caption{\small BERT text classification performance on CoLA and SST-2 datasets. NdLinear improves accuracy and ROC AUC with $\approx$85\% fewer parameters in the classification head.}
\label{tab:text_classify}
\begin{tabular}{ccccc}
\toprule
{Dataset} & {Method} & {Params (Head)} & {Accuracy} & {ROC AUC} \\ \midrule
\multirow{2}{*}{CoLA}  
    & Linear & 1,544 & 0.7790 $\pm$ 0.0143 & 0.7127 $\pm$ 0.0264 \\ 
    & NdLinear & 222 & \textbf{0.7906 $\pm$ 0.0142} & \textbf{0.7405 $\pm$ 0.0209} \\ \midrule
\multirow{2}{*}{SST-2} 
    & Linear & 1,544 & 0.8872 $\pm$ 0.0079 & 0.8867 $\pm$ 0.0080 \\ 
    & NdLinear & 222 & \textbf{0.8933 $\pm$ 0.0093} & \textbf{0.8932 $\pm$ 0.0073} \\ 
\bottomrule
\end{tabular}
\end{table}

\subsection{Time Series}
\label{appen:time_series}

\textbf{Time Series Forecasting. } In our experiments using RNNs, we used a sequence length of 24 and a forecast horizon of 12 for all models. The models were trained for 100 epochs using the Adam optimizer with a learning rate of 0.02 and a batch size of 128. The dataset was split into training, validation, and evaluation sets with proportions of 60\%, 20\%, and 20\%, respectively. We set the hidden size to 96, used a single recurrent layer, and applied a dropout rate of 0.3.

For Transformer-based Forecasting tasks, the experiments were conducted using a time series transformer model with a model dimension and hidden dimension both set to 32, a single transformer layer, and a dropout rate of 0.1. The GELU activation function was employed throughout, and the models were trained for 10 epochs with a batch size of 128 and a learning rate of 0.001. All models optimized using Adam and mean squared error as the loss function.

\textbf{Time Series Classification. } We set the hidden size of all RNN layers to 128 and used 3 recurrent layers, with a batch size of 32 and a learning rate of 0.005. Models were trained for 200 epochs using the Adam optimizer and cross-entropy loss.

\subsection{Tabular Data}
\label{appen:tabular}
We compare the performance of the Linear and NdLinear models. The Linear Model uses two linear layers for feature extraction, while the NdLinear Model replaces them with NdLinear layers.

\textbf{Classification. } Target labels were one-hot encoded.
The Linear Model utilized fully connected layers with input dimension $[11]$ and hidden dimension $[128]$, followed by ReLU and a final linear output layer. The NdLinear Model used custom NdLinear layers with input dimensions $[11, 1]$ and hidden dimensions $[11, 64]$, also followed by ReLU and a final linear output layer. 
Both models are trained over 40 epochs with a batch size of 32 and a learning rate of $0.0001$ using AdamW optimizer. Data was randomly shuffled and split into 80\% training and 20\% testing sets. Cross-entropy loss is used for training, and classification accuracy is used for model evaluation.

\textbf{Regression. } Target labels were kept as continuous values. 
The Linear Model utilized fully connected layers with input dimension $[14]$ and hidden dimension $[128]$, followed by ReLU and a final linear output layer. The NdLinear Model used custom NdLinear layers with input dimensions $[14, 1]$ and hidden dimensions $[32, 64]$, also followed by ReLU and a final linear output layer. 
Both models are trained over 40 epochs with a batch size of 32 and a learning rate of $0.0002$ using AdamW optimizer. Data was randomly shuffled and split into 80\% training and 20\% testing sets. MSE loss is used for both training and model evaluation.

\begin{table}[!ht]
  \centering
  \caption{\small NdLinear with MLPs on tabular datasets. For classification (Cardio Disease), the metric is Accuracy (higher is better). For regression (Delivery Time), the metric is MSE (lower is better).}
  \label{tab:ts_tf_classify_final}
  \begin{tabular}{@{}cllcc@{}} %
    \toprule
    Dataset & Task & Method & \#Params & Metric \\ 
    \midrule
    \multirow{2}{*}{Cardio Disease}
        & \multirow{2}{*}{\shortstack{Classif.\\(Accuracy)}} & Linear   & 18\,306 & 0.7265 \\
        &                            & NdLinear & 5\,962  & \textbf{0.7321} \\ 
    \midrule
    \multirow{2}{*}{Delivery Time}
        & \multirow{2}{*}{\shortstack{Regress.\\(MSE)}} & Linear   & 18\,561 & 70.508 \\ 
        &                            & NdLinear & 7\,873  & \textbf{67.824} \\ 
    \bottomrule
  \end{tabular}%
\end{table}

\subsection{Vision}
\label{appen:vision}
\textbf{Image Classification with CNN. } The NdLinear version uses three transforms with hidden dimensions of $(64,8,8)$, while the Linear version uses a single hidden dimension of $256$. Models were trained for 50 epochs using Adam optimizer (learning rate $0.001$), batch size 64, cross-entropy loss, and mixed-precision training on CUDA when available. 

\begin{table}[h]
\centering
\caption{\small Image classification with CNNs on CIFAR-10 (top-1 Acc.) and CIFAR-100 (top-5 Acc.). NdLinear achieves higher accuracy with fewer parameters.}
\label{tab:cnn-image-classify}
\begin{tabular}{cccc}
\toprule
{Dataset} & {\#Params} & {Method} & {Accuracy} \\ \midrule
\multirow{2}{*}{CIFAR-10} 
    & 1.07M & Linear & 0.7426 $\pm$ 0.0025 \\ 
    & 65k & NdLinear & \textbf{0.7689 $\pm$ 0.0060} \\ \midrule
\multirow{2}{*}{CIFAR-100} 
    & 1.09M & Linear & 0.6587 $\pm$ 0.0075 \\ 
    & 433k & NdLinear & \textbf{0.7096 $\pm$ 0.0121} \\ \bottomrule
\end{tabular}
\end{table}

\textbf{Vision Transformers (ViT) \cite{dosovitskiy2021image}. } Training used a batch size of 512, AdamW optimizer, learning rate $2.75 \times 10^{-4}$ for 30 epochs, and a distillation temperature of 3. Input images ($224 \times 224$) were augmented with random cropping and horizontal flipping.
\begin{figure}[h] %
    \centering
    \begin{subfigure}[b]{0.48\columnwidth} %
    \centering
        \includegraphics[width=\linewidth]{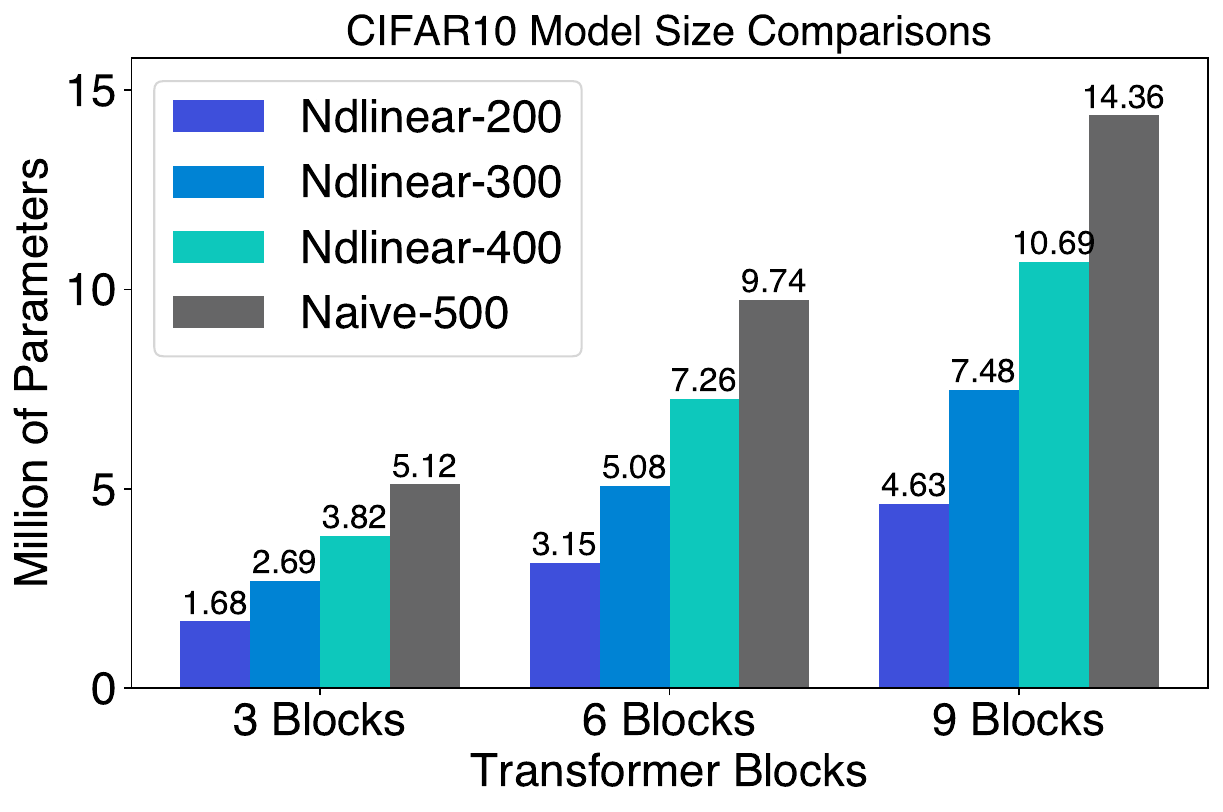}
    \end{subfigure}
    \hfill 
    \begin{subfigure}[b]{0.48\columnwidth} %
        \centering
        \includegraphics[width=\linewidth]{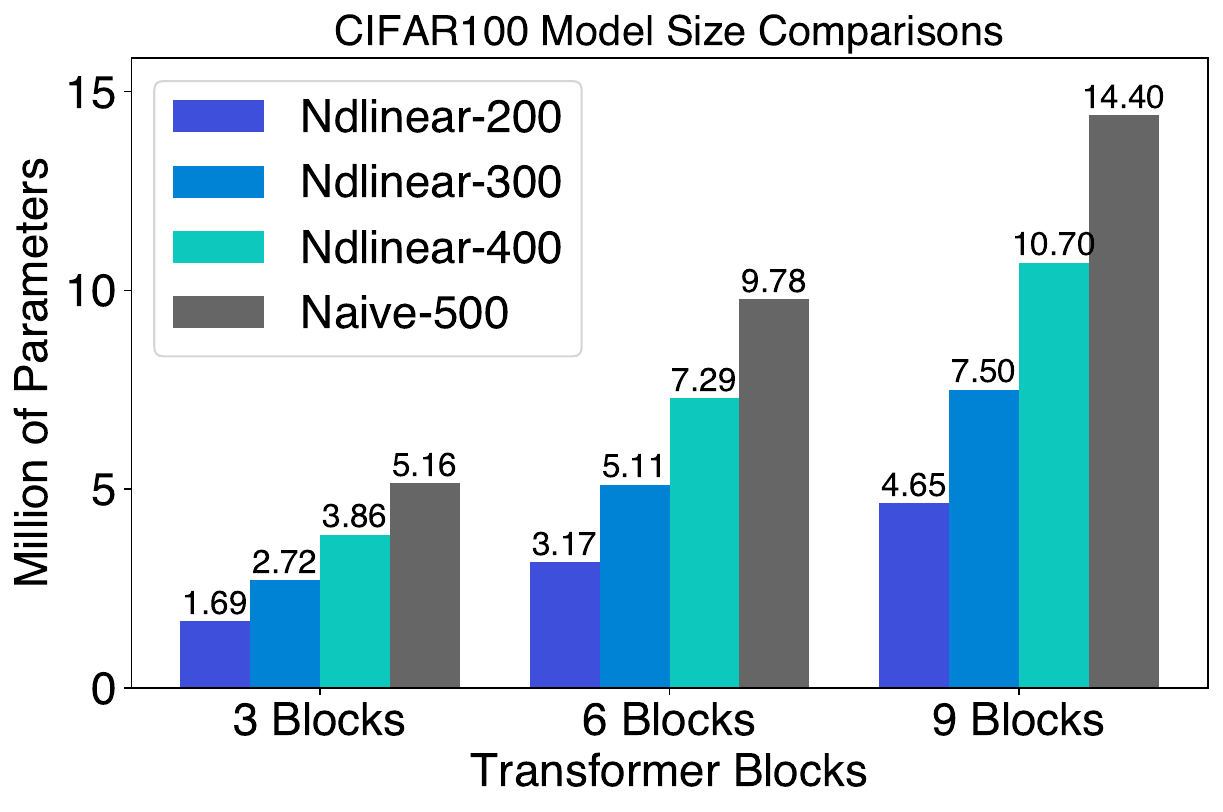}
    \end{subfigure}
    \caption{\small NdLinear's efficiency. Reduced ViT model parameter counts on CIFAR-10 and CIFAR-100 for a distillation task. }
    \label{fig:new_combined_vit_size}
\end{figure}

\begin{table}[H]
\centering
\caption{\small NdViT vs. Naive ViT: Accuracy and parameter efficiency on CIFAR-10 (Acc@1) and CIFAR-100 (Acc@5). NdViT shows improved accuracy with fewer parameters.}
\label{tab:vit-scaling}
\begin{tabular}{cccccc}
\toprule
\multirow{2}{*}{\begin{tabular}[c]{@{}c@{}}\centering {Dataset}\end{tabular}} & \multirow{2}{*}{\begin{tabular}[c]{@{}c@{}} Num. of \\ Transformers\end{tabular}} & \multicolumn{3}{c}{{NdViT (Ours)}} & {Naive} \\ \cmidrule(lr){3-6} 
& & \multicolumn{1}{c}{{200}} & \multicolumn{1}{c}{{300}} & {400} & {500} \\ \midrule
\multirow{3}{*}{\begin{tabular}[c]{@{}c@{}}\centering {CIFAR10}\end{tabular}} & {3 Blocks} & \multicolumn{1}{c}{65.77 $\pm$ 0.47} & \multicolumn{1}{c}{67.53 $\pm$ 0.70} & \textbf{69.00 $\pm$ 1.27} & 62.09 $\pm$ 0.40 \\ \cmidrule(lr){2-6} 
& {6 Blocks} & \multicolumn{1}{c}{68.48 $\pm$ 0.75} & \multicolumn{1}{c}{70.20 $\pm$ 0.73} & \textbf{72.03 $\pm$ 0.46} & 65.19 $\pm$ 0.64 \\ \cmidrule(lr){2-6} 
& {9 Blocks} & \multicolumn{1}{c}{70.27 $\pm$ 0.35} & \multicolumn{1}{c}{71.50 $\pm$ 0.58} & \textbf{72.53 $\pm$ 0.54} & 68.52 $\pm$ 1.24 \\ \midrule
\multirow{3}{*}{\begin{tabular}[c]{@{}c@{}}\centering {CIFAR100}\end{tabular}} & {3 Blocks} & \multicolumn{1}{c}{70.78 $\pm$ 1.36} & \multicolumn{1}{c}{73.10 $\pm$ 1.06} & \textbf{74.14 $\pm$ 1.66} & 69.34 $\pm$ 0.88 \\ \cmidrule(lr){2-6} 
& {6 Blocks} & \multicolumn{1}{c}{73.60 $\pm$ 0.83} & \multicolumn{1}{c}{75.07 $\pm$ 0.14} & \textbf{76.37 $\pm$ 0.71} & 73.84 $\pm$ 0.39 \\ \cmidrule(lr){2-6} 
& {9 Blocks} & \multicolumn{1}{c}{74.24 $\pm$ 0.32} & \multicolumn{1}{c}{75.52 $\pm$ 0.73} & \textbf{76.61 $\pm$ 0.26} & {75.60 $\pm$ 0.70} \\ \bottomrule
\end{tabular}
\end{table}
\textbf{Diffusion Transformers (DiT) \cite{peebles2023scalable}. } We used a learning rate of $1 \times 10^{-4}$ for $256 \times 256$ images, mixed-precision (\texttt{bfloat16}) training, automatic gradient accumulation, and a batch size of 256, varying model depth and attention heads.
\begin{table}[h]
    \centering
    \caption{\small FID-10k scores for DiT models: Pre-trained vs. NdLinear variants with fewer parameters.}
    \begin{tabular}{lccc}
        \toprule
        & {NdLinear} & {NdLinear} & {Baseline} \\
        \midrule
        {Parameter Count} & 619M & 563M & 674M \\
        {FID-10k} & 5.4876 & 5.9420 & 5.4109 \\
        \bottomrule
    \end{tabular}
    \label{tab:fid_scores}
\end{table}

\begin{figure}
    \begin{subfigure}[b]{0.3\columnwidth} %
        \centering
        \includegraphics[width=\linewidth, page=1]{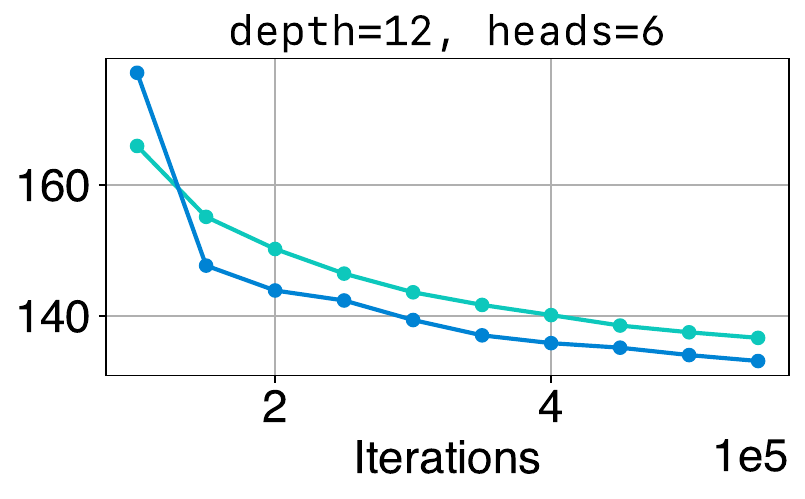}
        \label{fig:dit_fid_page1}
    \end{subfigure}
    \hfill
    \begin{subfigure}[b]{0.3\columnwidth} %
        \centering
        \includegraphics[width=\linewidth, page=2]{exp_plots/dit-scratch.pdf}
        \label{fig:dit_fid_page2}
    \end{subfigure}
    \hfill
    \begin{subfigure}[b]{0.3\columnwidth} %
        \centering
        \includegraphics[width=\linewidth, page=3]{exp_plots/dit-scratch.pdf}
        \label{fig:dit_fid_page3}
    \end{subfigure}
    \caption{DiT achieving lower (better) FID scores for image generation on ImageNet-100 when trained from scratch with comparable parameters. }
    \label{fig:new_combined_efficiency_showcase}
\end{figure}

\section{Detailed Ablation Studies}
\label{app:ablations}

\subsection{Per-Mode Bias Impact}
\label{app:bias}

We ablate per-mode bias terms across widths on the Radius Bump task. 

We generate $1{,}000$ train and $200$ test samples. Hidden widths \{16, 32, 64, 128\}. We have NdLinear-MLPs replacing Linear blocks with NdLinear, using per-axis hidden shapes \{(4,4), (8,8), (16,16), (32,32)\} matched to the hidden widths above. For all models, training uses Adam (learning rate $10^{-3}$) for $4{,}000$ epochs under identical schedules, with early stopping when training loss $< 10^{-4}$; loss is mean squared error (MSE). We report parameter count and test MSE (mean $\pm$ std).

The benefit of per-mode bias grows with width, reaching \textbf{+15.2\%} MSE improvement at width 128.

\begin{table}[h]
\centering
\caption{\small Per-mode bias ablation on Radius Bump}
\begin{tabular}{ccccc}
\toprule
\textbf{Width} & \textbf{Bias} & \textbf{MSE (mean)} & \textbf{MSE (std)} & \textbf{Params} \\
\midrule
16  & False & 0.00332264 & 0.00173532 & 36  \\
16  & True  & 0.00335201 & 0.00157289 & 46  \\
32  & False & 0.00332793 & 0.00170700 & 72  \\
32  & True  & 0.00318228 & 0.00163146 & 90  \\
64  & False & 0.00336938 & 0.00160766 & 144 \\
64  & True  & 0.00305152 & 0.00145630 & 178 \\
128 & False & 0.00339010 & 0.00173152 & 288 \\
128 & True  & 0.00287480 & 0.00139560 & 354 \\
\bottomrule
\end{tabular}
\end{table}

\subsection{Axis Ordering Sensitivity}
\label{app:ordering}

We permute axes (original, reverse, random) and measure CIFAR-100 performance retention. Finding: robustness to ordering ($\leq$ 4 pp spread).

\begin{table}[h]
\centering
\caption{\small Axis ordering sensitivity (CIFAR-100)}
\begin{tabular}{lc}
\toprule
\textbf{Variant (axes)} & \textbf{Accuracy vs. baseline} \\
\midrule
Original order & \textbf{100\%} \\
Reverse        & 99\% $\pm$ 1\% \\
Random         & 96\% $\pm$ 1\% \\
\bottomrule
\end{tabular}
\end{table}

\subsection{Radius Bump: Last-Layer Entropy, Performance, and Compression}
\label{app:entropy}

We evaluate Dense MLP vs.\ NdLinear on the Radius Bump task across three difficulty levels (shell thickness $\sigma \in \{0.10, 0.20, 0.30\}$). Inputs are $x \in [-1,1]^{10}$ with i.i.d.\ coordinates $x_i \sim \mathrm{U}(-1,1)$; the target is
$
y = \exp\!\bigl(-(\|x\|-0.8)^2/(2\sigma^2)\bigr).
$
For each $\sigma$ we generate $1{,}000$ train and $200$ test samples. Architectures: (i) Dense MLP baselines with hidden widths \{16, 32, 64, 128\} and depths \{2, 3\}; (ii) NdLinear-MLPs replacing Linear blocks with NdLinear, using per-axis hidden shapes \{(4,4), (8,8), (16,16), (32,32)\} matched to the dense widths above and depths \{2, 3\}. For all models, training uses Adam (learning rate $10^{-3}$) for $4{,}000$ epochs under identical schedules, with early stopping when training loss $< 10^{-4}$; loss is mean squared error (MSE). We report parameter count, forward-pass FLOPs (from analytical op counts), test MSE (mean $\pm$ std), and last-layer average output entropy (AveEntropy).

\begin{table}[h]
\centering
\caption{\small Radius Bump ($\sigma=0.10$; thin / hardest). Lower AveEntropy indicates more compact last-layer representations.}
\label{tab:radius-010}
\resizebox{\columnwidth}{!}{%
\begin{tabular}{r l r c r l l r r}
\toprule
\textbf{Width} & \textbf{Kind} & \textbf{Depth} & \textbf{Hidden Shape} & \textbf{Params} & \textbf{Train MSE} & \textbf{Test MSE $\pm$ Std} & \textbf{FLOPs ($\cdot 10^9$)} & \textbf{AveEntropy} \\
\midrule
16  & dense & 2 & (16)     & 193    & 1.17e-5 & 0.00086 ± 0.00176 & 2.78   & 0.476 \\
16  & dense & 3 & (16)     & 465    & 1.37e-6 & 0.00087 ± 0.00189 & 6.70   & 0.625 \\
16  & nd    & 2 & (4, 4)   & 46     & 1.19e-5 & 0.00086 ± 0.00186 & 0.66   & 0.405 \\
16  & nd    & 3 & (4, 4)   & 86     & 1.18e-5 & 0.00086 ± 0.00186 & 1.24   & 0.518 \\
32  & dense & 2 & (32)     & 385    & 1.18e-5 & 0.00086 ± 0.00169 & 5.54   & 0.459 \\
32  & dense & 3 & (32)     & 1,441  & 2.93e-6 & 0.00084 ± 0.00188 & 20.75  & 0.560 \\
32  & nd    & 2 & (8, 8)   & 90     & 1.20e-5 & 0.00086 ± 0.00182 & 1.30   & 0.321 \\
32  & nd    & 3 & (8, 8)   & 234    & 1.16e-5 & 0.00086 ± 0.00178 & 3.37   & 0.426 \\
64  & dense & 2 & (64)     & 769    & 1.19e-5 & 0.00086 ± 0.00169 & 11.07  & 0.407 \\
64  & dense & 3 & (64)     & 4,929  & 1.59e-8 & 0.00085 ± 0.00196 & 70.98  & 0.507 \\
64  & nd    & 2 & (16, 16) & 178    & 1.12e-5 & 0.00086 ± 0.00175 & 2.56   & 0.292 \\
64  & nd    & 3 & (16, 16) & 722    & 1.12e-5 & 0.00086 ± 0.00180 & 10.40  & 0.377 \\
128 & dense & 2 & (128)    & 1,537  & 5.79e-6 & 0.00089 ± 0.00253 & 22.13  & 0.363 \\
128 & dense & 3 & (128)    & 18,049 & 1.84e-9 & 0.00089 ± 0.00158 & 259.91 & 0.469 \\
128 & nd    & 2 & (32, 32) & 354    & 1.12e-5 & 0.00086 ± 0.00158 & 5.10   & 0.295 \\
128 & nd    & 3 & (32, 32) & 2,466  & 8.71e-6 & 0.00086 ± 0.00204 & 35.51  & 0.286 \\
\bottomrule
\end{tabular}
}
\end{table}

\begin{table}[h]
\centering
\caption{\small Radius Bump ($\sigma=0.20$; medium). NdLinear often attains both lower error and lower last-layer entropy.}
\label{tab:radius-020}
\resizebox{\columnwidth}{!}{%
\begin{tabular}{r l r c r l l r r}
\toprule
\textbf{Width} & \textbf{Kind} & \textbf{Depth} & \textbf{Hidden Shape} & \textbf{Params} & \textbf{Train MSE} & \textbf{Test MSE $\pm$ Std} & \textbf{FLOPs ($\cdot 10^9$)} & \textbf{AveEntropy} \\
\midrule
16  & dense & 2 & (16)     & 193    & 7.75e-4 & 0.00256 ± 0.00235 & 2.78   & 0.493 \\
16  & dense & 3 & (16)     & 465    & 1.29e-5 & 0.00166 ± 0.00118 & 6.70   & 0.592 \\
16  & nd    & 2 & (4, 4)   & 46     & 9.91e-4 & 0.00273 ± 0.00273 & 0.66   & 0.410 \\
16  & nd    & 3 & (4, 4)   & 86     & 1.01e-3 & 0.00275 ± 0.00272 & 1.24   & 0.521 \\
32  & dense & 2 & (32)     & 385    & 6.03e-4 & 0.00266 ± 0.00234 & 5.54   & 0.461 \\
32  & dense & 3 & (32)     & 1,441  & 5.92e-6 & 0.00180 ± 0.00117 & 20.75  & 0.560 \\
32  & nd    & 2 & (8, 8)   & 90     & 9.92e-4 & 0.00273 ± 0.00232 & 1.30   & 0.323 \\
32  & nd    & 3 & (8, 8)   & 234    & 1.19e-4 & 0.00192 ± 0.00089 & 3.37   & 0.428 \\
64  & dense & 2 & (64)     & 769    & 3.17e-4 & 0.00368 ± 0.00344 & 11.07  & 0.403 \\
64  & dense & 3 & (64)     & 4,929  & 1.84e-8 & 0.00128 ± 0.00190 & 70.98  & 0.518 \\
64  & nd    & 2 & (16, 16) & 178    & 8.35e-4 & 0.00252 ± 0.00216 & 2.56   & 0.293 \\
64  & nd    & 3 & (16, 16) & 722    & 9.56e-6 & 0.00071 ± 0.00019 & 10.40  & 0.344 \\
128 & dense & 2 & (128)    & 1,537  & 2.53e-5 & 0.00454 ± 0.00330 & 22.13  & 0.364 \\
128 & dense & 3 & (128)    & 18,049 & 2.01e-10 & 0.00159 ± 0.00102 & 259.91 & 0.471 \\
128 & nd    & 2 & (32, 32) & 354    & 6.07e-4 & 0.00233 ± 0.00211 & 5.10   & 0.296 \\
128 & nd    & 3 & (32, 32) & 2,466  & 2.33e-6 & 0.00081 ± 0.00019 & 35.51  & 0.284 \\
\bottomrule
\end{tabular}
}
\end{table}

\begin{table}[h]
\centering
\caption{\small Radius Bump ($\sigma=0.30$; thick / easiest). NdLinear depth-3 variants achieve the lowest errors with the lowest entropies.}
\label{tab:radius-030}
\resizebox{\columnwidth}{!}{%
\begin{tabular}{r l r c r l l r r}
\toprule
\textbf{Width} & \textbf{Kind} & \textbf{Depth} & \textbf{Hidden Shape} & \textbf{Params} & \textbf{Train MSE} & \textbf{Test MSE $\pm$ Std} & \textbf{FLOPs ($\cdot 10^9$)} & \textbf{AveEntropy} \\
\midrule
16  & dense & 2 & (16)     & 193    & 3.65e-3 & 0.00716 ± 0.00157 & 2.78   & 0.505 \\
16  & dense & 3 & (16)     & 465    & 1.87e-4 & 0.00210 ± 0.00084 & 6.70   & 0.627 \\
16  & nd    & 2 & (4, 4)   & 46     & 5.91e-3 & 0.00827 ± 0.00261 & 0.66   & 0.406 \\
16  & nd    & 3 & (4, 4)   & 86     & 2.73e-3 & 0.00710 ± 0.00345 & 1.24   & 0.527 \\
32  & dense & 2 & (32)     & 385    & 2.03e-3 & 0.00636 ± 0.00209 & 5.54   & 0.461 \\
32  & dense & 3 & (32)     & 1,441  & 1.42e-5 & 0.00166 ± 0.00063 & 20.75  & 0.588 \\
32  & nd    & 2 & (8, 8)   & 90     & 3.53e-3 & 0.00564 ± 0.00190 & 1.30   & 0.323 \\
32  & nd    & 3 & (8, 8)   & 234    & 6.78e-5 & 0.00045 ± 0.00021 & 3.37   & 0.424 \\
64  & dense & 2 & (64)     & 769    & 8.76e-4 & 0.00968 ± 0.00175 & 11.07  & 0.406 \\
64  & dense & 3 & (64)     & 4,929  & 1.34e-5 & 0.00152 ± 0.00029 & 70.98  & 0.521 \\
64  & nd    & 2 & (16, 16) & 178    & 3.01e-3 & 0.00502 ± 0.00142 & 2.56   & 0.293 \\
64  & nd    & 3 & (16, 16) & 722    & 9.46e-6 & 0.00029 ± 0.00021 & 10.40  & 0.341 \\
128 & dense & 2 & (128)    & 1,537  & 1.66e-4 & 0.01057 ± 0.00324 & 22.13  & 0.362 \\
128 & dense & 3 & (128)    & 18,049 & 2.03e-10 & 0.00249 ± 0.00051 & 259.91 & 0.474 \\
128 & nd    & 2 & (32, 32) & 354    & 1.78e-3 & 0.00460 ± 0.00096 & 5.10   & 0.296 \\
128 & nd    & 3 & (32, 32) & 2,466  & 3.41e-5 & 0.00041 ± 0.00014 & 35.51  & 0.278 \\
\bottomrule
\end{tabular}
}
\end{table}

\begin{figure}
    \centering
    \includegraphics[width=0.5\linewidth]{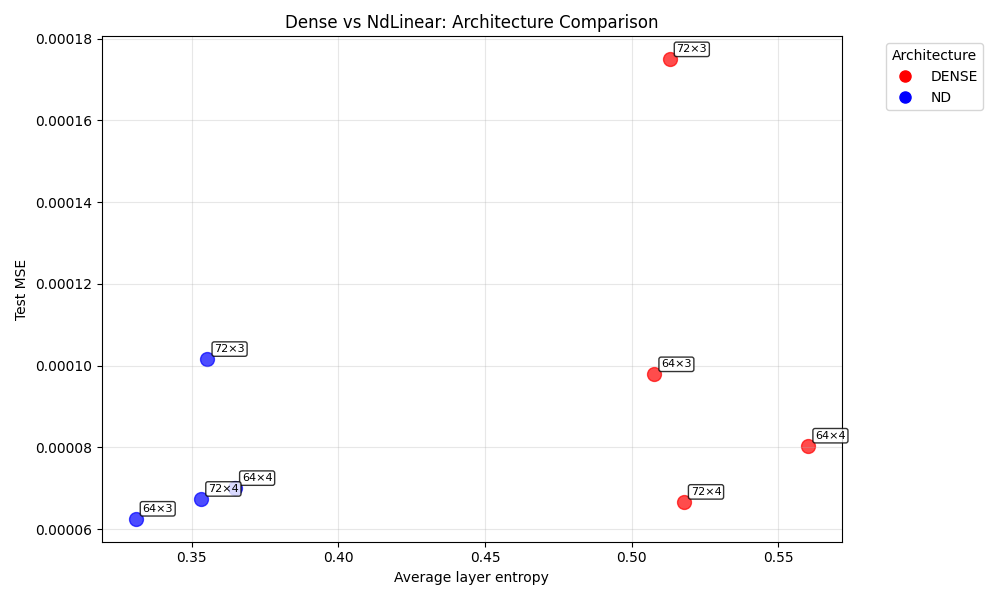}
    \caption{\small $\alpha = 0.1$ (hard): Narrow bump, very challenging.}
    \label{fig:placeholder2}
\end{figure}

\begin{figure}
    \centering
    \includegraphics[width=0.5\linewidth]{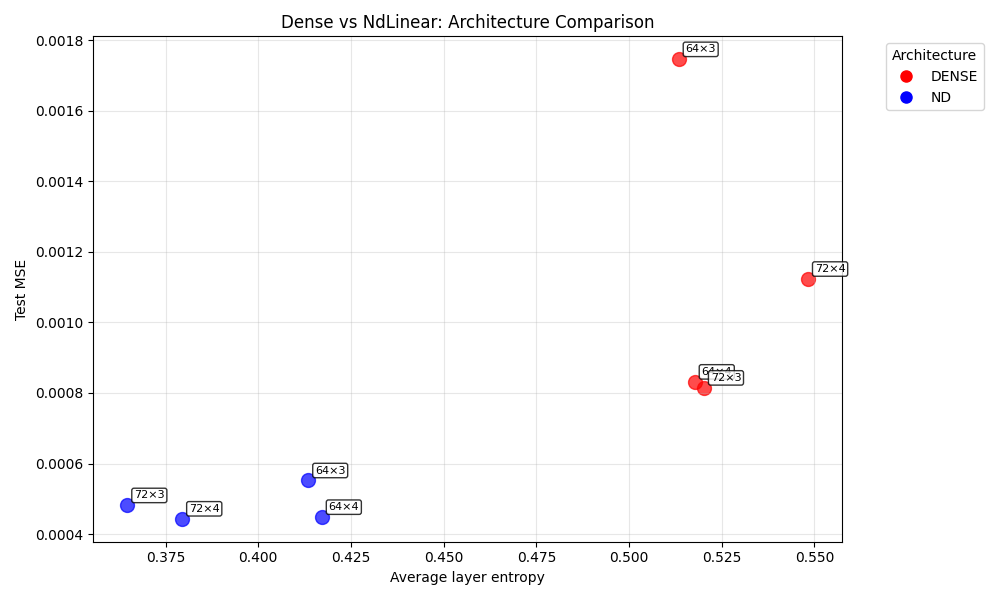}
    \caption{\small $\alpha = 0.2$ (medium): Moderate difficulty.}
    \label{fig:placeholder1}
\end{figure}
\begin{figure}
    \centering
    \includegraphics[width=0.5\linewidth]{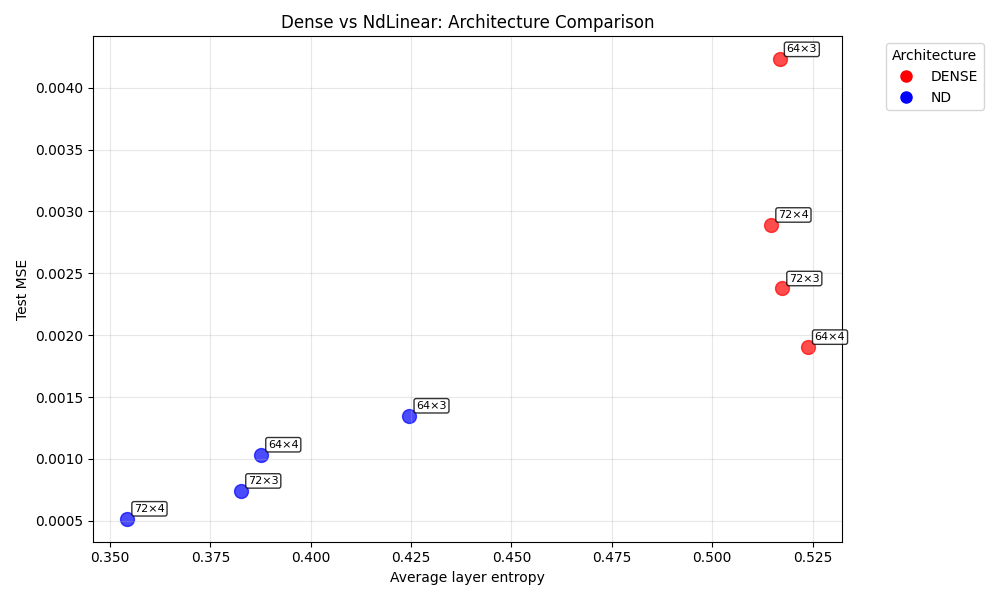}
    \caption{\small $\alpha$ = 0.3 (easy): Wide bump, easier to approximate.}
    \label{fig:placeholder3}
\end{figure}

Across all $\sigma$, NdLinear variants consistently exhibit lower AveEntropy than parameter-matched dense models at similar widths/depths, while using markedly fewer parameters and FLOPs. At $\sigma=0.20$ and $\sigma=0.30$, depth-3 NdLinear achieves the \emph{lowest} test errors (e.g., $0.00071 \pm 0.00019$ at width 64, and $0.00029 \pm 0.00021$ at width 64) together with \emph{lower} entropies (0.344 and 0.341, respectively) than dense counterparts (0.00128 $\pm$ 0.00190 with entropy 0.518; 0.00152 $\pm$ 0.00029 with entropy 0.521). For the hardest setting ($\sigma=0.10$), NdLinear matches dense test error while maintaining \emph{lower entropy} and large reductions in parameters and FLOPs. These patterns support the claim that lower last-layer entropy paired with equal-or-better error corresponds to better compression.

\subsection{Hyperparameter Sensitivity}
\label{app:hyperparam}

Grid search over learning rate, hidden size, and batch size on CIFAR-100. Finding: stable efficiency metrics and competitive accuracy across settings.

\begin{table}[h]
\centering
\caption{\small Hyperparameter sensitivity sweep (CIFAR-100)}
\small
\begin{tabular}{ccccccc}
\toprule
\textbf{LR} & \textbf{Hidden Size} & \textbf{Batch} & \textbf{GFLOPs} & \textbf{Latency (s)} & \textbf{Params} & \textbf{Acc@5 (\%)} \\
\midrule
0.001 & 128,4,4 & 128 & 0.848 & 0.001415 & 232,876 & 72.75 \\
0.001 & 256,2,2 &  64 & 0.901 & 0.001451 & 138,760 & 74.30 \\
0.001 & 128,4,4 &  64 & 0.848 & 0.001381 & 232,876 & 72.67 \\
0.001 & 128,4,4 &  32 & 0.848 & 0.001409 & 232,876 & 72.48 \\
0.001 & 256,2,2 &  32 & 0.901 & 0.001388 & 138,760 & 73.80 \\
0.001 & 256,2,2 & 128 & 0.901 & 0.001400 & 138,760 & \textbf{74.75} \\
0.01  & 128,4,4 & 128 & 0.848 & 0.001318 & 232,876 & 71.62 \\
0.01  & 128,4,4 &  64 & 0.848 & 0.001376 & 232,876 & 66.73 \\
0.01  & 128,4,4 &  32 & 0.848 & 0.001364 & 232,876 & 66.56 \\
0.01  & 256,2,2 &  64 & 0.901 & 0.001257 & 138,760 & 64.86 \\
0.01  & 256,2,2 & 128 & 0.901 & 0.001312 & 138,760 & 70.20 \\
0.01  & 256,2,2 &  32 & 0.901 & 0.001323 & 138,760 & 63.75 \\
\bottomrule
\end{tabular}
\end{table}
\vspace{-0.3em}
Note: GPU memory usage constant at 34.04-34.40 MB across all configurations.

\subsection{Sample Efficiency}
\label{app:sample}

Fix task structure and vary data size; measure samples needed to reach target error at different entanglement levels.

\begin{itemize}[topsep=2pt,itemsep=2pt,parsep=0pt]
\item \textbf{Nearly-separable tasks} ($\alpha = 0.1$): NdLinear reached target MSE with only \textbf{2,000} samples; parameter-matched linear model required \textbf{over 10,000} samples.
\item \textbf{Highly-entangled tasks} ($\alpha = 0.9$): Standard linear model achieved target MSE with \textbf{15,000} samples; NdLinear struggled to match this performance even with \textbf{25,000} samples.
\end{itemize}

\subsection{Training and Memory Overhead}
\label{app:overhead}

Measure peak activation memory and per-epoch training time after replacing a single dense GEMM with mode-wise GEMMs. Finding: empirical overheads are small ($<3\%$ memory, $<2\%$ time) as seen in Table \ref{tab:overhead}.

\begin{table}[h]
\centering
\caption{\small Empirical overheads across architectures}
\begin{tabular}{lcc}
\toprule
\textbf{Model} & \textbf{Peak Mem (MB)} & \textbf{Epoch Time (s)} \\
\midrule
CIFAR-100 CNN & 35.17 $\rightarrow$ 36.91 (+2.0\%) & 47.2 $\rightarrow$ 47.8 (+0.6\%) \\
ETTh1 RNN     & 32.58 $\rightarrow$ 33.41 (+1.2\%) & 12.3 $\rightarrow$ 12.6 (+1.2\%) \\
Vision Transformer & 127.3 $\rightarrow$ 130.1 (+1.1\%) & 179 $\rightarrow$ 185 (+1.6\%) \\
\bottomrule
\end{tabular}
\label{tab:overhead}
\end{table}

\subsection{Comparison with Alternative Structured Layers}
\label{app:structured}

CNN head on CIFAR-100 comparing NdLinear to TRL/TCL and TT. Finding: NdLinear achieves higher Acc@5 with fewer params, lower FLOPs, and lower latency.

\begin{table}[h]
\centering
\caption{\small NdLinear vs. TRL/TCL vs. TT (CIFAR-100 Acc@5)}
\begin{tabular}{lccccc}
\toprule
\textbf{Method} & \textbf{Mem (MB)} & \textbf{Acc@5} & \textbf{Latency (s)} & \textbf{FLOPs (G)} & \textbf{Params} \\
\midrule
\textbf{NdLinear} & \textbf{35.16} & \textbf{0.7133} & \textbf{0.000976} & \textbf{0.843} & \textbf{433,588} \\
TRL/TCL           & 35.60          & 0.6935          & 0.001116          & 3.97            & 548,032 \\
TT                & 100.44         & 0.5617          & 0.005871          & 5.25            & 769,316 \\
\bottomrule
\end{tabular}
\label{tab:trl_tcl2}

\end{table}

\end{document}